\documentclass[10pt,journal,compsoc]{IEEEtran}

\usepackage{latexsym}
\usepackage{amsmath}
\usepackage{amsfonts}
\usepackage{helvet}
\usepackage{courier}
\usepackage{url}
\usepackage{float,color}
\usepackage{times}
\usepackage{algorithm}
\usepackage{algorithmic}
\usepackage{epsfig}
\usepackage{amssymb}
\usepackage{flushend}
\usepackage{stfloats}
\usepackage{amsmath}
\usepackage{graphicx,subfigure}
\usepackage{url}
\usepackage{epstopdf}
\usepackage[none]{hyphenat}
\usepackage{multicol}
\usepackage{tocstyle}
\usepackage{dsfont,amsfonts,amssymb,amsmath,color}
\usepackage{multirow}
\usepackage{booktabs}

\newtocstyle{compact}{%
  \settocfeature[1]{entryhook}{\bfseries}%
  \settocfeature[1]{entryvskip}{2pt plus 5pt}%
  \settocfeature[1]{leaders}{\hfill}%
}
\usetocstyle{compact}

\def\c{{\bf c}}

\def\e{{\bf e}}

\def\w{{\bf w}}

\def\x{{\bf x}}

\def\0{{\bf 0}}
\def\1{{\bf 1}}

\def\CM{{\mathcal C}}

\def\RB{{\mathbb R}}

\def\ph{\mbox{\boldmath$\phi$\unboldmath}}

\def\etal{{\em et al.\/}\,}

\newcommand{\nop}[1]{}

\newcommand{\eg}{{e.g.}}

\newtheorem{definition}{Definition}

\begin{document}

\title{Machine Learning with World Knowledge: The Position and Survey}

\author{Yangqiu Song and Dan Roth
	\IEEEcompsocitemizethanks{\IEEEcompsocthanksitem Yangqiu Song is with the
		Department of Computer Science and Engineering,
		Hong Kong University of Science and Technology,
		Clear Water Bay, Hong Kong. \protect
		E-mail: yqsong@cse.ust.hk
		\IEEEcompsocthanksitem Dan Roth is with the
		Department of Computer Science,
		University of Illinois at Urbana-Champaign,
		Urbana, IL 21218 USA. \protect
		E-mail: danr@illinois.edu
	}
}

\markboth{Yangqiu Song and Dan Roth: Machine Learning with World Knowledge}%
{Yangqiu Song and Dan Roth: Machine Learning with World Knowledge}

\IEEEtitleabstractindextext{%
\begin{abstract}
	Machine learning has become pervasive in multiple domains, impacting a wide variety of applications, such as knowledge discovery and data mining, natural language processing, information retrieval, computer vision, social and health informatics, ubiquitous computing, etc.
	Two essential problems of machine learning are how to generate features and how to acquire labels for machines to learn.
	Particularly, labeling large amount of data for each domain-specific problem can be very time consuming and costly.
	It has become a key obstacle in making learning protocols realistic in applications.
	In this paper, we will discuss how to use the existing general-purpose world knowledge to enhance machine learning processes, by enriching the features or reducing the labeling work.
	We start from the comparison of world knowledge with domain-specific knowledge, and then introduce three key problems in using world knowledge in learning processes, i.e., explicit and implicit feature representation, inference for knowledge linking and disambiguation, and learning with direct or indirect supervision.
	Finally we discuss the future directions of this research topic.
\end{abstract}

}

\maketitle

\tableofcontents

\section{Introduction}
\label{sec:introduction}

Machine learning has become pervasive in multiple domains, impacting a wide variety of applications, such as knowledge discovery and data mining, natural language processing, information retrieval, computer vision, social and health informatics, ubiquitous computing, etc.
Two major problems of machine learning in practice are how to generate or extract features from data and how to acquire labels for machines to learn.
There have been many studies about feature engineering and labeling work reduction in the past decades.
\begin{itemize}
	\item {\bf Feature extraction and representation}. Feature engineering, such as handcrafting features for domain dependent problems, has been recognized as a key problem in applications (such as Kaggle\footnote{\url{https://www.kaggle.com/Competitions}} or KDD Cup\footnote{\url{http://www.sigkdd.org/kddcup/index.php}}).
	Given the features, with or without labels, one can perform feature selection~\cite{LiuFSK1998} or feature extraction~\cite{SametFMM2005,saul06spectral,Cunningham2015} to find a better representation than the handcrafted features for learning algorithms. More recently, deep learning has been proposed to deal with big data through end-to-end learning which enables the representation learning within the deep architecture of neural networks~\cite{HinSal06,YannLeCun15}. However, there are still problems related to high-level intelligence that current machine learning systems cannot handle. For example, human knowledge about the world is highly structured. When human imagines something, the higher-order relationships among ``knowledge dots'' (facts, entities, events or activities, etc.) can be the key clue. However, current machine learning systems may not be able to capture the inference process of such remote relationship among the dots. Even with the rapid development of deep learning, which has much better representation ability of data, discovering relationships is still a problem. For example, although current neural network language model can capture long-short term memory of words~\cite{Mikolov132,HochreiterS97,SukhbaatarSWF15} in language process, it is still difficult to capture the global dependencies across documents, e.g., cross-document co-reference~\cite{SinghSPM11}.
	\item {\bf Labeling work reduction}. Labeling large amount of data for each domain-specific problem can be very time consuming and costly. It has become a key obstacle in making learning protocols realistic in applications. Machine learning community has also elaborated to reduce the labeling work done by human for supervised machine learning algorithms or to improve unsupervised learning with only minimum supervision. For example, semi-supervised learning~\cite{ChaSchZie06} is proposed to use only partially labeled data and a lot of unlabeled data to perform learning with the hope that it can perform as good as fully supervised learning. Transfer learning~\cite{Pan2010} uses the labeled data from other relevant domains to help the learning task in the target domain or learns multiple domains simultaneously. Both semi-supervised learning and transfer learning needs domain knowledge, and there are multiple ways to achieve these learning settings. However, there is no general solution or a principle when applying both learning settings to most tasks. In other words, for each of the target domain, specific domain knowledge is still needed to be engineered into the learning process. Crowdsourcing~\cite{Lease11,GuoWSH14} has been considered to acquire cheap labels from general-level human intelligence. However, current crowdsourcing mechanisms can still be applied to relatively simple and well-defined tasks, and it is still a challenge for applying machine learning to the labels for more diverse and more specific data~\cite{Lease11}.
\end{itemize}

We use some specific open problems from natural language processing and computer vision to further illustrate the above problems.

{\bf Example 1: text semantics and topics.}

Text semantic similarity/relatedness is one of the fundamental problem in natural language processing.
Regarding to different levels of text span, e.g., word, phrase, sentence, or document, there are different ways to compute the similarity/relatedness~\cite{jurgens-pilehvar:2015:EMNLP-Tutorials}.
For example, we consider short texts such as following~\cite{Wang2015IWK}.

{\it Text 1.1: On Feb. 10, 2007, Obama announced his candidacy for President of the United States in front of the Old State Capitol located in Springfield, Illinois.}

{\it Text 1.2: Bush portrayed himself as a compassionate conservative, implying he was more suitable than other Republicans to go to lead the United States.}

An intuitive way is to compute the bag-of-words similarity between two texts.
However, given the above text fragments, the overlapped words are few.
Nonetheless, the similarity/relatedness between these two fragments should be high, since they are talking to the same topic.
Therefore, one can consider to use the context of each word to enrich the similarity between two fragments, where the context could be obtained from a lot of texts from all over the world~\cite{Collobert11,TurianRaBe10Clean,Mikolov132,MikolovYZ13}.
Another way is to relate the words or entities in the texts by the external knowledge~\cite{Schuhmacher2014,WangChenguangICDM2015}.
For example, we know that ``Obama'' is related to ``Bush'' since they were both the President of the ``United States.''
Thus, if we can find a path between ``Obama'' and ``Bush'' in the external knowledge base, we can directly relate the two text fragments without seeing other words related to them.
Both approaches are not dependent to the target two pieces of short texts but leverage the knowledge from general purpose texts or knowledge bases.


{\bf Example 2: events: language and vision.}
Event extraction is another key component in natural language understanding.
Given its complex definition of event trigger, agents, instruments, targets, location, and time~\cite{NIST05,mitamura2015event}, a joint inference must be applied to identify the corresponding events.
Traditional event extraction approaches train the machine learning models based on the annotation on specific domains, e.g., 33 event types in ACE 2005~\cite{NIST05} or 38 event types in TAC KBP 2015~\cite{mitamura2015event}.
Consequently, the supervised learning systems easily overfit these domains.
However, there are many more types of events.
When generalizing the trained models to other domains, more annotation should be used.
Especially, the relationships among agents, instruments, and targets are very difficult to discover using the small number of annotated data.
Thus, a more global approach is expected to avoid training models overfitting small domains.
For example, the determination of event nugget can be decided by simply computing a structured similarity between seed examples and the new events~\cite{PengSR16}, where the similarity is coming from general purpose knowledge base, and the structure can be obtained by general purpose semantic role labeling~\cite{PunyakanokRY08}.
Furthermore, if we can extract knowledge about entities and relations from existing knowledge bases, such information can help us perform joint inference about the entities involved in the event extraction problem across documents.

Similarly in computer vision, event~\cite{TuMLCZ14,QiWuVTT2015} and scene~\cite{JohnsonKSLSBL15,KrishnaZGJHKCKL16} recognition problems are also the most key and complicated problems of image understanding.
When parsing an image, not only the pixels or features, but all of the functionality, attribute, intentionality, and causality factors should be considered.\footnote{\url{http://www.stat.ucla.edu/~sczhu/research_blog.html}}
The above constrains again make event/scene recognition problem a joint inference problem.
When performing inference, commonsense knowledge is one of the key problem to perform joint inference.
For example, when we see a fish, most likely it is in water.
When it is not in water, the scene may be interpreted as some specific event, e.g., fish hunting or cooking.

{\bf Example 3: co-reference.} Co-reference resolution is a problem of finding different entities or mentions in texts referring to the same person or thing. It is also one of the key problems in natural language understanding.
Typical co-reference resolution systems use rule-based method~\cite{Lee2013DCR} or learning-based method~\cite{ChangSR13}.
For the learning based methods, it is common to extract the features for the entities or mentions, and define some deterministic function to compare pairs of entities or mentions to identify whether they are co-referent.
Even if declarative constraints have been considered based on some background knowledge of co-reference in natural language~\cite{ChangSR13}, some of the hard problems are still very difficult for these systems to solve. For example, here are some examples of hard co-reference problem~\cite{Rahman2011CRW,Levesque11,PengKhRo15}.

{\it Example 3.1:} {\it [Martha Stewart]$_{e}$ is hoping people don't run out on her. [The celebrity]$_{NP}$ indicted on charges stemming from...}

{\it Example 3.2:} {\it [Martha Stewart]$_{e}$ is hoping people don't run out on {[Tom]$_{e}$}. [The celebrity]$_{NP}$ indicted on charges stemming from...}

{\it Example 3.3:} {\it [A bird]$_{e1}$ perched on the [limb]$_{e2}$ and [it]$_{pro}$ bent.}

Example 3.1 was initially shown in~\cite{Rahman2011CRW}. It illustrates that if a noun phrase (NP) refers to a named entity (e), some external knowledge about the entity as a celebrity can improve the determination of co-reference. Otherwise, only based on the lexical or syntactical features, the developed rules may not be perfect to generalize to other cases~\cite{Rahman2011CRW}.
Example 3.2 shows an even harder case. If the first sentence has two named entities, we should know which one is the celebrity to perform the inference.
Example 3.3 shows another example from the Winograd schema challenge~\cite{Levesque11,RahmanN12} indicating that we should have the commonsense knowledge that a bird cannot be bent whereas a branch of a tree can.
So here we have shown that certain knowledge about the entities, categories, or attributes can help identify the co-reference.

All the above examples show that traditionally when we perform machine learning, we mostly focused on how to train a model that can avoids overfitting and have best generalization ability.
However, even we have the best model and the parameter tuning skills, the machine learning algorithms may still lack of knowledge and the higher-order relationships about the entities they have seen, or they are still easily overfitting to a specific domain they are trained based on.
Therefore, more general approaches should be considered.

In this paper, we present the idea of ``machine learning with world knowledge.''
Instead of only considering the data in a specific domain, we also consider the general purpose knowledge about the world.
The general knowledge includes common and commonsense knowledge, and partially the domain dependent knowledge.
We position the idea of using world knowledge as an intersection of many fields, including machine learning, data mining, natural language processing, knowledge representation, etc.
We start with comparing the traditionally used domain/background knowledge for machine learning algorithms and the world knowledge.
Then we discuss why world knowledge is useful and what are the important problems when using world knowledge for machine learning algorithms.
Specifically, we need to adapt world knowledge, which is domain independent, to domain problems.
Then we introduce how the two important factors in machine learning algorithms, features and labels, are affected by world knowledge.
There are multiple ways to represent world knowledge as features for machine learning algorithms.
We survey the existing approaches and summarize them into three categories, homogeneous and heterogeneous explicit features and  implicit features.
To use world knowledge as supervision, we introduce the linking and inference techniques that can relate a domain problem to a general-purpose knowledge base.
Then we introduce some new learning paradigms that can be enabled by world knowledge.
Finally we discuss the future directions of the ideas of machine learning with world knowledge and conclude our paper.
Note that, we will not focus on machine learning for world knowledge acquisition or organization.
Instead, we assume that the world knowledge is already existing for machines to use.

\section{Domain vs. World Knowledge}
In this section, we present the concepts of domain knowledge and world knowledge, and extend to the related problems when using both of them.

\subsection{Domain Knowledge and Domain Adaptation}
As we mentioned in the introduction, most of the feature engineering needs domain knowledge.
For example, to identify a disease, certain related symptoms should be observed.
Moreover, supervised learning and semi-supervised learning~\cite{ChaSchZie06} both require labels or side information, e.g., must-link and cannot-link constraints~\cite{Basu08}, for the domain to perform machine learning.
The domain knowledge is reflected by the labels or the constraints.
In this section, we focus on three aspects of machine learning with domain knowledge.
First, we introduce the most intuitive semi-supervised learning based on partially labeled data, which is formulated as a generative process setting.
This is the simplest case of applying domain knowledge to machine learning algorithms, and strongly related to posterior regularization in Section~\ref{sec:constraintlearning}.
For more examples and learning settings of semi-supervised learning, we refer to the corresponding references for further reading~\cite{ChaSchZie06,Basu08}.
Then, we survey the declarative knowledge constrained learning since the declarative knowledge is very related to the form of world knowledge.
Finally, we focus on the domain knowledge that can be transferred to other domains.
On one hand, this learning setting is a very good comparison with semi-supervised learning to have more insight about domain knowledge.
On the other hand, we can compare it with setting of domain adaptation with world knowledge.

\subsubsection{Generative Semi-supervised Learning}
\label{sec:semisupervisedlearning}
A typical generative semi-supervised learning setting (e.g.,~\cite{NigamMTM00}) is to learn a set of parameters $\Theta$, given a set of unlabeled data $\{ {\mathcal X}, {\mathcal Y} \}$ and a small portion of labeled data $\{ {\mathcal X}_L, {\mathcal Y}_L \}$.
Then the maximum likelihood estimation of $\Theta$ is:
\begin{equation}\label{eq:likelihood}
\max_{\Theta} \log \sum_{\mathcal Y}P({\mathcal X}, {\mathcal Y} | \Theta) +
\log P({\mathcal X}_L, {\mathcal Y}_L | \Theta).
\end{equation}
This cannot be solved directly since there is a sum operation inside logarithm.
When the posterior of ${\mathcal Y}$ is analytically tractable, an expectation-maximization algorithm can be employed.
We can re-write the first term in the likelihood as:
\begin{align}\label{eq:factorizedlikelihood}
\max_{\Theta} \sum_{\mathcal Y} Q({\mathcal Y}) \log \frac{ P({\mathcal X}, {\mathcal Y} | \Theta) }{Q({\mathcal Y})} - \sum_{\mathcal Y} Q({\mathcal Y}) \log \frac{ P({\mathcal Y} | {\mathcal X} , \Theta) }{Q({\mathcal Y})} \nonumber \\
=\max_{\Theta} \sum_{\mathcal Y} Q({\mathcal Y}) \log \frac{ P({\mathcal X}, {\mathcal Y} | \Theta) }{Q({\mathcal Y})} + {\rm KL} [Q({\mathcal Y}) || P({\mathcal Y} | {\mathcal X} , \Theta)].
\end{align}
By substituting Eq.~(\ref{eq:factorizedlikelihood}) into Eq.~(\ref{eq:likelihood}), we can derive the EM algorithm.
In the E-step, we have $Q({\mathcal Y}) = P({\mathcal Y} | {\mathcal X},  \Theta)$, when there is analytical solution, to minimize the KL-divergence ${\rm KL} [Q({\mathcal Y}) || P({\mathcal Y} | {\mathcal X}, \Theta)]$.
In the M-step, we substitute $Q({\mathcal Y}) = P({\mathcal Y} | {\mathcal X},  \Theta^{old})$ to the first term and solve the following problem:
\begin{equation}\label{eq:mstep}
\Theta^{new} = \max_{\Theta} \sum_{\mathcal Y} Q({\mathcal Y}) \log { P({\mathcal X}, {\mathcal Y} | \Theta) } + \log P({\mathcal X}_L, {\mathcal Y}_L | \Theta),
\end{equation}
which can be solved by either analytics or gradient descent based algorithms.

This example of semi-supervised learning applies domain knowledge of partial labels to the generative learning process.
The parameter estimation is affected by the labeled set according to Eq.~(\ref{eq:mstep}).
However, the strong assumption is that the labeled data $\{ {\mathcal X}_L, {\mathcal Y}_L \}$ should be i.i.d. as the unlabeled data $\{ {\mathcal X}, {\mathcal Y} \}$.

\subsubsection{Declarative Constraints}
\label{sec:constraintlearning}
Declarative constraints are introduced in the context of using background knowledge to improve machine learning algorithms' performance.
For example, in natural language processing, one example is that in part-of-speech recognition, if we know that there should be one verb and one noun in a sentence, then we can constrain the unlabeled sentences to satisfy the linguistic knowledge~\cite{ganchev2010posterior}.
Another example is that in information extraction applied to citation domain, we can constrain that the word ``pp.'' corresponds to the ``page'' label~\cite{ChangRR12}.
Then even if there is no training data showing that ``pp.'' can be extracted as page information, we can still obtain this from the constraints.
In general, we can incorporate such constraints in a declarative way so that they can be formulated as certain logic forms that machines can read and use.
Representative studies include the constrained conditional models (CCM)~\cite{ChangRR12}, generalized expectation criteria~\cite{MannM10}, 
measurements in Bayesian framework~\cite{LiangJK09}, and posterior regularization~\cite{ganchev2010posterior,HuMLHX16}.
For example, in posterior regularization, the constraints are used to limit the solution region that the parameters can fall in.
Specifically, originally, in the E-step, we solve the $Q({\mathcal Y})$ to minimize the KL-divergence ${\rm KL} [Q({\mathcal Y}) || P({\mathcal Y} | {\mathcal X}, \Theta)]$.
In posterior regularization, we solve the following constrained optimization problem in E-step:
\begin{eqnarray}
\min_{Q({\mathcal Y}), \mbox{\boldmath$\xi$} }{\rm KL} [Q({\mathcal Y}) || P({\mathcal Y} | {\mathcal X}, \Theta)] + \lambda ||\mbox{\boldmath$\xi$}||_{\beta} \nonumber\\
s.t. \;\; {\rm E}_{Q({\mathcal Y})}[\phi ({\mathcal Y}, {\mathcal X})]- {\bf b} \leq \mbox{\boldmath$\xi$},
\end{eqnarray}
where $\phi ({\mathcal Y}, {\mathcal X})$ and ${\bf b}$ are the variables to make constraints satisfy certain formulations~\cite{ganchev2010posterior}, $\mbox{\boldmath$\xi$}$ are the slack variables to allow the constraints being violated, and $\lambda$ is the weight of Langrange to penalize the overall errors that the constraints can be made.
Posterior regularization can be seen as variational approximation of both generalized expectation criteria~\cite{MannM10} and measurements in Bayesian framework~\cite{LiangJK09}.
Some of the must-link and cannot-link based clustering algorithms~\cite{LuL04,basu2004probabilistic,song2013constrained} can also be categorized in to this learning paradigm, even though they did not relate themselves to this more general learning framework.
If we inspect the EM algorithm they used, we can simply find that their E-step is constrained, while M-step remains the same as the traditional EM algorithms used for clustering.
Compared to these algorithms, the semi-supervised learning introduced in Section~\ref{sec:semisupervisedlearning} follows the Eq.~(\ref{eq:mstep}) to modify the M-steps in the EM algorithms while keep the E-step as the traditional EM algorithm used for generative models~\cite{NigamMTM00}.

When we only focus on the MAP (maximum a posteriori) estimation of the parameters, the above formulation degenerates to the CCM framework~\cite{ChangRR12}.
There are two advantages that CCM introduced.
First, CCM can be used to decouple the learning and inference parts of the algorithm.
The learning part corresponds to the parameter estimation of the learning model, while the inference part corresponds to the label assignment for the structured output of learning algorithm.
We will introduce in more detail in Section~\ref{sec:overview}.
Then this decoupling will further introduce the second advantage.
If we can focus only on the inference part, we can use a much simpler representation and algorithm to handle the constraints.
For example, we can in general represent satisfiability of boolean formulas (SAT) as a linear algebra form and solve the SAT problem using some well developed optimization tools to solve it~\cite{FangzhenLin20103}.
For example, in CCM, it uses integer linear programming (ILP) and A-star algorithms to solve the problems~\cite{ChangRR12}.
It has been shown that this general form of constraints is useful to represent many constrained learning problems~\cite{ChangRR12}.

\subsubsection{Transfer Learning}\label{sec:transferlearning}
Transfer learning~\cite{Torrey2009,Pan2010} is a learning paradigm that uses data in relevant tasks to help the target machine learning tasks.
Formally, if we have a source domain data ${\mathcal S}=\{ {\mathcal X}_S, {\mathcal Y}_S \}=\{{\bf x}_{S_i}, {\bf y}_{S_i}\}_{i=1}^{N_S}$, where ${\bf x}_{S_i}$ is the feature vector of sample $i$ in source domain, ${\bf y}_{S_i}$ is the label vector (without loss of generality we use vector to represent label(s) of a data sample), and $N_S$ is the number of available data in source domain.
We also have some target domain data ${\mathcal T}=\{ {\mathcal X}_T, {\mathcal Y}_T \}=\{{\bf x}_{T_i}, {\bf y}_{T_i}\}_{i=1}^{N_T}$, where ${\bf x}_{T_i}$ is the feature vector of sample $i$ in target domain, ${\bf y}_{T_i}$ is the label vector, and $N_T$ is the number of available data in source domain.
In most cases, we have $N_T\ll N_S$. Sometimes, we have no labeled data but some unlabeled data in the target domain.
The goal of transfer learning is to use the source domain data ${\mathcal S}$ to help improve the learning/prediction performance of the target domain data ${\mathcal T}$.
In the problem of transfer learning, the two domains can be different in terms of ${\mathcal X}_T \neq {\mathcal X}_S$, ${\mathcal Y}_T \neq {\mathcal Y}_S$ or $P({\mathcal S}) \neq P({\mathcal T})$.
For example, we can train a newsgroup classifier based on ``Christian vs. Hockey'' and transfer the knowledge of classification to ``Atheism vs. Autos''~\cite{RainaNK06}, where ${\mathcal Y}_T \neq {\mathcal Y}_S$ and $P({\mathcal X}_T) \neq P({\mathcal X}_S)$.
We can use a ``motorbike'' object detector to detect ``bicycle'' from images~\cite{AytarZ11}, where $P({\mathcal X}_T) \neq P({\mathcal X}_S)$.
We can also even transfer the knowledge from text to images~\cite{YangCXDY09}, where ${\mathcal X}_T \neq {\mathcal X}_S$.

The domain here in transfer learning corresponds to the specific tasks that are relevant to the target task.
The domain knowledge usually means the implicit knowledge incorporated in the learned models, the distributions of source data, or the latent factors that are related to the factors in the target domain~\cite{Pan2010}.
Thus, when applying the knowledge from source domain to the target domain, we first need to know what are the relevant tasks that can provide such knowledge.
Second, we need to develop specific algorithm that can incorporate or use the existing knowledge from the source domain.
These can be regarded as the major characteristics of domain knowledge in transfer learning.

\subsection{World Knowledge and Domain Adaptation}

If we analyze the above learning paradigms, we can find something in common: they all use domain knowledge to help learning algorithms better find a solution for a domain specific problem.
For example, semi-supervised learning needs seed labels or must-link and cannot-link constraints for the domain, which should be i.i.d. with the unlabeled data and the new prediction data.
Declarative constraint driven learning needs the background knowledge about the task, and incorporates the knowledge into the constraints.
Transfer learning requires the source task relevant to the target task, and the knowledge can be transferred or adapted.
This means that when dealing with a new task, all of them need human to evaluate the new task and incorporate the ``correct'' knowledge in to the learning process.
Compared to the above learning paradigms, the paradigm ``machine learning with world knowledge'' does not require human to justify the domain knowledge.
Whereas, it uses the general-purpose knowledge, which can be obtained from large scale general knowledge base, or in general the data from the Web, to help learning algorithms to improve the learning performance.

In the past decades, especially in recent years, there are a lot of general-purpose knowledge bases (or knowledge graphs) developed, e.g., WordNet~\cite{wordnet}, Cyc project~\cite{researchCyc}, Wikipedia, Freebase~\cite{freebase}, KnowItAll~\cite{knowitall}, TextRunner~\cite{BankoCSBE07}, WikiTaxonomy~\cite{PonzettoS07}, Probase~\cite{wu2011taxonomy}, DBpedia~\cite{auer2007dbpedia}, YAGO~\cite{suchanek2007yago}, NELL~\cite{NELL},
Illinois-Profiler~\cite{Zhiye2015} and Knowledge Vault~\cite{dong2014knowledge}.
We call these knowledge bases world knowledge~\cite{GabrilovichM05}, because they are universal knowledge that are either collaboratively annotated by human labelers or automatically extracted from big data.
For example, collaboratively constructed knowledge bases include WordNet, Cyc, Wikipedia, and Freebase.
Knowledge bases extracted based on information extraction includes Probase, DBpedia, YAGO, NELL, Illinois-Profiler, and Knowledge Vault.
A more comprehensive of comparison of scales and methodologies of knowledge bases can be found in~\cite{Nickel0TG16}.
When world knowledge is annotated or extracted, it is not collected for any specific domain.
The first paper explicitly mentioning machine learning with world knowledge is~\cite{GabrilovichM05}.

In general, slightly different from traditional definition~\cite{GabrilovichM05}, we summarize world knowledge as commonsense knowledge, common knowledge, and domain knowledge following~\cite{CambriaSWH11}, since we find this way will better distinguish different kinds of knowledge that can be used.
\begin{itemize}
	\item Commonsense knowledge.\footnote{\url{https://en.wikipedia.org/wiki/Common_sense}} Commonsense knowledge is an very important sub-topic in artificial intelligence~\cite{RussellAIM}. Here we refer commonsense knowledge as the knowledge that an ordinary person is expected to know, but they normally leave unstated when they write or talk. For example, ``cats can hunt mice;'' ``birds can fly;'' etc. Thus, commonsense is the most difficult part of knowledge to collect from the Web since there is too few resources mentioning such knowledge in purpose.
	\item Common knowledge.\footnote{\url{https://en.wikipedia.org/wiki/Common_knowledge}} Common knowledge refers to the knowledge that humans generally know about the world. For example, ``the United States is a country;'' ``the current President of the United States;'' etc. Different from commonsense knowledge, there is a lot of resources mentioning such knowledge on the Web. Note that people with different educational or cultural background should have different common knowledge. Nonetheless, collectively speaking, common knowledge can be extracted mostly from the Web now (in a noisy way).
	\item Domain knowledge. Domain knowledge is the knowledge in a specific domain. For example, the meaning of a term in molecular biology may only be understood by a biologist. Currently, some world knowledge bases such as Wikipedia also contain some of the domain knowledge. However, for a complete ontology or dictionary of concepts, a more domain specific knowledge base is expected.
\end{itemize}
From the above categorization, we can see that most of the mentioned world knowledge bases only tried to cover the common knowledge part, and partially cover the commonsense and domain knowledge,
especially the knowledge bases constructed based on information extraction.
Therefore, it is still far away from solving every problem using current world knowledge bases. However, the common knowledge is already very useful for us to enrich our data representation and introduce weak supervision.
In general, we will have machine learning algorithms to learn from following data:
\begin{equation}\label{eq:likelihoodW1}
\max_{\Theta} \log P({\mathcal Y} , {\mathcal X}_{W}, \Theta)
\end{equation}
or
\begin{equation}\label{eq:likelihoodW2}
\max_{\Theta} \log P({\mathcal Y}_{W} , {\mathcal X}, \Theta),
\end{equation}
where ${\mathcal X}_{W}$ are the features that can be obtained from the world to extend the representation of data and ${\mathcal Y}_{W}$ can be weak labels automatically obtained from the world knowledge base. Then the learned model parameter $\Theta$ should be able to apply to new coming data:
\begin{equation}\label{eq:classificationW}
y^* = \arg\max_{y\in \mathcal{Y}} P(y|{{\bf x}, \Theta}).
\end{equation}
Thus, from Eq.~(\ref{eq:classificationW}) we can see that, we will have a domain adaptation problem to either adapt the world knowledge about the labels to the target domain, or adapt the data in feature space to the target domain.

In the following of the paper, we will introduce how to leverage the existing world knowledge for machines to learn.

\section{Machine Learning with World Knowledge: An Overview}
\label{sec:overview}

In both collaborative data collection (or crowdsourcing)~\cite{chilton2013cascade,bragg2013crowdsourcing,MengT0C15,Sun2015hierarchical} and information extraction from the Web~\cite{knowitall,BankoCSBE07,NELL}, machine learning is widely used.
Moreover, learning algorithms are also used to do inference over knowledge bases~\cite{NiepertD15,Nickel0TG16}.
Instead of using machine learning to construct knowledge bases or do inference over knowledge bases,
``machine learning with world knowledge'' considers how to use the existing world knowledge bases to help improving existing learning algorithms or applications.

In this section, we consider the general machine learning framework that can incorporate world knowledge into machine learning algorithms.
There are multiple ways to use world knowledge for machine learning.
As we mentioned in the introduction, two key machine learning problems are feature extraction/representation and label reduction.
Thus, the intuitive ways to incorporate world knowledge into machine learning algorithms can be classified into these two categories.
However, world knowledge is not designed for any specific domain.
For example, if the world knowledge is about all kinds of named entities in the world, then when we want to process the documents about entertainment or sports, the world knowledge about names of celebrities and athletes may help while the terms used in science and technology may not be very useful.
Thus, another key issue is how we should specify the world knowledge to the domain specific tasks, or adapt the world knowledge to domains.
Thus, here we summarize the issues about machine learning with world knowledge into three categories: representation, inference, and learning, which is analogous to other machine learning problems such as sequence labeling~\cite{ChangRR12} and Bayesian networks~\cite{KollerFriedman09}.
We summarize the three categories of problems in Table~\ref{tab:learningworldknowledge}.

\begin{table*}[t]
	\centering
	{\small
		\caption{Machine learning with world knowledge problems analogous  to sequence labeling and Bayesian networks.}
		\begin{tabular}{p{0.12\textwidth}|p{0.27\textwidth}p{0.27\textwidth}p{0.26\textwidth}}
			\toprule
			& Representation & Inference & Learning \\
			\midrule
			Sequence Text Labeling              &     Distributional/distributed representations.   & Sequence label assignment; decoder. & Parameter estimation; weight learning. \\
			\hline
			Bayesian Network         & Distribution family; Independence.  &  Variable elimination; belief propagation; variational inference; Sampling. & Parameter estimation; network structure learning. \\
			\hline
			Learning with World Knowledge         & Knowledge as features; graphs; labels.  & Knowledge specification (entity linking; semantic parsing; disambiguation). & Parameter estimation; domain adaptation. \\
			\bottomrule
		\end{tabular}
	}
	\label{tab:learningworldknowledge}
\end{table*}

{\bf Representation.} As many other machine learning algorithms, learning with world knowledge needs the representation of data samples. For example, in sequence labeling for text, such as named entity recognition, to predict each word's label, a set of features should be extracted. The features can be one-hot distributional lexical features~\cite{BrownPdLM92,KotlermanDSZ10}, or neural network word embeddings~\cite{BengioDVJ03,Mikolov132,MikolovYZ13}. In Bayesian networks, we need to determine which distribution should be used to describe the data, and the (conditional) independency among random variables~\cite{WainwrightJ08,KollerFriedman09}. Since world knowledge is usually about the entities all over the world and their relations, the representation of knowledge can be categorial and structured. Moreover, the knowledge can be either used as features or used as (indirect) labels. Thus, the representation of world knowledge used for machine learning can be multiple ways.

{\bf Inference.} Inference means to infer more knowledge from the data, or discover the relationships from data. For example, in sequence labeling, label assignment is determined by considering all possible assignment of labels in a sequence. However, there are more efficient ways to do this, e.g., beam search~\cite{TillmannN03} or Viterbi algorithm~\cite{Rabiner1989}. Other possible ways such as A-star algorithm~\cite{ChangRR12} or policy based search~\cite{DaumeLM09} an also be applied.
In Bayesian networks, inference involves to infer the posteriors or marginal of random variables such as using variational inference, belief propagation, and random sampling~\cite{WainwrightJ08,KollerFriedman09}.
Here in learning with world knowledge, we consider the inference problem as specifying world knowledge to domain problems (For the problem of inferring more knowledge given a knowledge base, such as link prediction, please refer to the corresponding survey paper~\cite{Nickel0TG16}). For example, given the knowledge base and the document, we want to infer the most probable categories of the entities in the document, and their relationships. The process of grounding the entities in a document to the knowledge base is usually called entity linking~\cite{RothJCC14,ShenWH15}. If the relations are also considered, it is usually called semantic parsing~\cite{Mooney07,ArtziFZ13}.
In general, we want to solve the ambiguity problem of the knowledge for a specific problem, by considering either the structural label relationship as sequence labeling problem or posterior inference as Bayesian network inference problems.

{\bf Learning.} Learning refers to the process that estimate the parameters of models.
For example, in sequence labeling, the parameters are the weights for the features.
In Bayesian networks, learning can refer to learning the parameters of the distributions, or learning the structure of latent variables.
Similar to other machine learning problems, learning with world knowledge also has a learning process.
Depending on different representations of world knowledge, the learning processes are also different.
Moreover, one particular issue is that world knowledge is not built for a specific domain.
Thus, the learning process will also have the problem of domain adaption similar to what we introduced in transfer learning.
However, in transfer learning, the domain adaptation usually refers to adapt the knowledge from one domain to another, where in learning with world knowledge, domain adaptation refers to adapt general knowledge to domains.

In the following sections, we will survey existing work and summarize them in the above three categories.

\section{Representation: World Knowledge as Features}
\label{sec:representation}
In this section, we survey the existing studies on using machine earning with world knowledge as features.
Particularly, we categorize the feature representations as explicit features, implicit features, and graph based features.


\subsection{Explicit Homogeneous Features}
Most of the traditional distributional representation can be regarded as explicit features, if the representation is generated based on a corpus in a specific domain.
For example, we can use the co-occurrence of syntactic or semantic patterns in the context~\cite{Hindle1990,Grefenstette92,Grishman1994,Faure98,FaureN99,cimiano2003}.
If the corpus is a world knowledge base, such as Wikipedia, then more specific distributed representations can be developed.
Here, we review the most significant development of knowledge base based representations for textual documents, i.e., the explicit semantic analysis (ESA)~\cite{GabrilovichM09}, probabilistic conceptualization (PC)~\cite{song2011short}, and their extension and combinations, since these models reveal important insight of generating features with world knowledge.
To summarize from the modeling perspective, analogous to the image conceptualization frameworks discussed in~\cite{Zhu03}, we introduce and analyze three ways to generate the representations: descriptive, generative and discriminative models.

For world knowledge representation, we consider generating world knowledge based features $\x_W=\c$ from the original features $\x=\{{\bf e}_1,...,{\bf e}_M\}$, where ${\bf e}_i$'s are the features related to a term (a word or an entity) in a document.\footnote{For other types of data, such image, the features for an entity could be different.}
In the descriptive and generative models, we consider to model the probability $P({\bf e}_1,...,{\bf e}_M | \c)$.
In the discriminative model, we consider directly modeling the probability $P(\c | \e_1,...,\e_M )$.
The major discussion of this section follows the previous paper~\cite{song2015short}.

\subsubsection{Explicit Semantic Analysis: Generative Models}

The first paper using the term ``world knowledge'' \cite{GabrilovichM05} extends the bag-of-words features with the categories in Open Directory Project (ODP), and shows that it can help improve text classification with additional knowledge.
Following this, by mapping the text to the semantic space provided by Wikipedia pages or other ontologies, it has been proven to be useful for short text classification~\cite{GabrilovichM06,GabrilovichM07,GabrilovichM09}, clustering~\cite{HuJian2008,Hu2009EIE,HuXiaohua2009,Fodeh2011}, and information retrieval~\cite{Egozi2011}.
In this line of research, the approaches are generally called Explicit Semantic Analysis (ESA).

ESA~\cite{GabrilovichM09} simply combines the weighted concepts of each term in a short text.
We use $\e_m = (e_{m,1},...,e_{m,T}) \in {\RB}^{T}_{+}$ to represent the concept vector of the term $e_m$.
For example, we can set $e_{m,t}=f(n(e_m, c_t))$ as a function of the co-occurrence of the term $e_m$ and $c_t$, where $e_m$ is an entity and $c_t$ is a higher-level concept.
In the original ESA, it uses TF-IDF (term frequency-inverse document frequency) score of $e_m$ shown in the $t$-th Wikipedia page, which is denoted as a concept $c_t$.
We use a vector $\c = (c_1, ..., c_T)\in {\RB}^{T}_{+}$ to denote the concept proportion that can describe the whole text containing $E=\{e_1, \ldots, e_M\}$.
Then ESA recalls the concepts with scores as this:
\begin{equation}
{\c} = \sum\nolimits_{m=1}^M w_m {\e}_m,
\end{equation}
where $w_m$ is the weight associated to $e_m$, e.g., the TF-IDF score of $e_m$ in the short text. The benefit of using this representation is that the values in the concept vectors $\e_m$ are not restricted to the co-occurrence frequencies, but can be arbitrarily tuned.

ESA can be regarded as a generative model since it uses the concept-term relationship as the evidence of generated features of terms, and estimates the latent concept distribution which generates the features.
If we formulate the probability  $P(\e_1,...,\e_M | \c)$ as:
\begin{align}\label{eq:ESA-prob}
P(\e_1,...,\e_M | \c) &= \prod\nolimits_{m=1}^M P(\e_m | \c) \\ \nonumber
&\propto \prod\nolimits_{m=1}^M\exp\{- || \e_m - \c ||^2 \},
\end{align}
where $P(\e_m | \c)$ is assumed to be a Gaussian distribution centered by the underlying concept distribution $\c$.
Then $\c = \frac{1}{M}\sum\nolimits_{m=1}^M \e_m$ is the maximum likelihood estimate with the probability $P(\e_1,...,\e_M | \c)$.
Here $P(\e_m | \c)$ is more flexible and not necessarily to be factorized as $\Pi_t^T P(e_m|c_t)$.
For example, $e_{m,t} \; (t=1,...,T)$ in the concept vector $\e_m$ can be the co-occurrence frequency of concept $c_t$ and term $e_m$ in the same sentence or same document. We can also define $e_{m,t} \triangleq P(c_t|e_m)$ which is the typicality of a concept $c_t$ to describe the term $e_m$, or $P(e_m|c_t )$, which is the typicality of how much a term $e_m$ can instantiate the concept $c_t$.

\subsubsection{Probabilistic Conceptualization: Descriptive Models}
Probabilistic conceptualization~\cite{song2011short} uses a different mechanism of getting concepts of each term/entity.
It has been applied to Twitter messages clustering~\cite{song2011short}, short text categorization~\cite{Wang14}, bag-of-words labeling~\cite{SunXW015}, search relevance measurement~\cite{Song14}, search log mining~\cite{Hua2013,WangZWMW15}, advertising keywords semantic matching~\cite{Liu2012,Kim2013}, and semantic frame identification~\cite{ParkHW16}.

Given a set of terms (words or multiple-word expressions, phrases, text segments, etc.) $E=\{e_1, \ldots, e_M\}$ in a short
text\footnote{Parsing short text to be words or multi-word expressions can be non-trivial~\cite{Song14}. We ignore this since it is not the focus of this paper.}, probabilistic conceptualization tries to find the concepts associated with scores that can best describe the terms.
Suppose we have a general and open domain concept set $\CM = \{c_1, \ldots, c_T\}$.
In probabilistic conceptualization, it makes the naive Bayes assumption of the conditional probabilities and uses
\begin{equation}\label{eq:entity-concept-posterior}
P(c_t | E) = {P(E|c_t) P(c_t)}/{P(E)}
\propto P(c_t) \prod\nolimits_{m=1}^{M} P(e_m|c_t)
\end{equation}
as the score associated with $c_t$.
Here, $P(e_m|c_t)=\frac{n(e_m, c_t)}{n(c_t)}$ where $n(e_m, c_t)$ is the co-occurrence frequency of concept $c_t$ and term $e_m$ in the sentences used by information extraction, and $n(c_t)$ is the overall number of concept $c_t$. Moreover, $P(c_t)=\frac{n(c_t)}{\sum_t n(c_t)}$ is normalized by the number of all the concepts in ${C}$.
The basic assumption behind this model is that given each concept $c_t$, all the observed terms $e_m \in E$ are conditionally independent. Then it uses the probability $P(c_t | E)$ to rank the concepts and selects the concepts with the largest probabilities to represent the text containing the terms in $E$.

The probabilistic conceptualization can be regarded as a simple causal Markov model, since it imposes the partial order of the probabilities of concept-term relationship.
We first assume the conditional independency of ${\bf e}_m$ given ${\bf c}$: $P({\bf e}_1,...,{\bf e}_M | \c) = \Pi_m^M P({\bf e}_m | \c)$.
Then we define $P({\bf e}_m | \c)\propto \Pi_t^T P({e_{m,t} | P(e_m | c_t))} =\Pi_t^T P(e_m | c_t)^{e_{m,t}}$ as a multinomial distribution where $P(e_m | c_t)$ is calculated based on the evidence of co-occurrence in knowledge base (explained under Eq. (\ref{eq:entity-concept-posterior})).
We define $e_{m,t} = 1$ if for this trial $c_t$ is selected as the description of the short text and $e_{m,t'}= 0$ for $t'\neq t$.
Now we can factorize $P({\bf e}_1,...,{\bf e}_M | \c)$ as:
\begin{equation}
P(e_1|c_1)^{e_{1,1}} \cdot ... \cdot P(e_1|c_T)^{e_{1,T}} \cdot ... \cdot P(e_M|c_T)^{e_{M,T}} ,
\end{equation}
By incorporating the prior $P(\c)\triangleq \prod_{t=1}^{T} P(c_t)$, we can re-write the posterior of $\c$:
\begin{align}\label{eq:entity-concept-total-posterior}
P(\c| \e_1,...,\e_M) &\propto P(\e_1,...,\e_M | \c) P(\c)  \\ \nonumber
& = \prod\nolimits_{t=1}^{T} P(c_t) \prod\nolimits_{m=1}^{M} P(e_m|c_t)^{e_{m,t}}.
\end{align}
Then selecting the top $k$ concepts using Eq.~(\ref{eq:entity-concept-posterior}) among all the $T$ concepts can be considered as the maximum a posterior (MAP) estimation of this posterior in Eq.~(\ref{eq:entity-concept-total-posterior}).
This illustrates what probabilistic conceptualization really optimizes.
Thus, if one of the probability $P(e_m|c_t)$ equals to zero, then the whole probability $P(\c| \e_1,...,\e_M)$ equals to zero.
Even if a smoothing technique can be applied~\cite{song2011short}, the probability mass $P(\c| \e_1,...,\e_M)$ could be too small to be reasonable in this case.

We can see that both the simple descriptive and generative approaches factorize the probability as $\prod\nolimits_{m=1}^M P(\e_m | \c)$, which do not consider the relationships between $e_m$'s.
Thus, a generative + descriptive model that tries to jointly model $P(\e_1,...,\e_M | \c)$ to incorporate the relationships between terms with more descriptive power is also introduced~\cite{song2015short}.

\subsubsection{Hierarchical Classification: Discriminative Model}
Another way for conceptualization is to classify the short text onto a predefined taxonomy or ontology~\cite{LiuYWZCM05,GopalY13,Cissetel01142046,SongR14,SongR15}.
Classification can be regarded as the discriminative model which wants to estimate $\c$ by directly modeling the probability $P(\c | \e_1,...,\e_M )$.
For example, we can learn (or simply find) a set of projection vectors $\w_t, t = 1,...,T$, to project the observed text to maximize $P(c_t | \w_t, \e_1,...,\e_M) = \frac{1}{Z} f(\w_t , g(\e_1,...,\e_M)) $, where the concept vector is considered as a feature vector to generate the representation of the short text. A typical $g(\e_1,...,\e_M)$ can be $\frac{1}{M}\sum\nolimits_{i=m}^M \e_m$.
Since hierarchical classification to an extremely large set of labels will be very costly, this may not be a best choice when we are trying to use world knowledge base simply as features for other machine learning tasks.

The summary of the above discussion is shown in Table~\ref{tab:representation}.

\begin{table*}[tbp]
	\centering
	\caption{Summarization of Learning with World Knowledge Features.}
	{\footnotesize
		\begin{tabular}{p{0.1\textwidth}|p{0.35\textwidth}|p{0.18\textwidth}|p{0.28\textwidth}}
			\toprule
			Representation & Explicit Flat Features &  Explicit Heterogeneous Features & Implicit Features\\
			\midrule
			Traditional Approaches        & Corpus driven distributional representations~\cite{Hindle1990,Grefenstette92,Grishman1994,Faure98,FaureN99,cimiano2003} &  Features~\cite{Fodeh2011} based on word sense disambiguation~\cite{BudanitskyH06} &  Topic models~\cite{deerwester1990indexing,Hofmann99,Blei03}, word embeddings and language models~\cite{BengioDVJ03,MorinB05,MnihH08,TurianRaBe10Clean,Collobert11,Mikolov133,Mikolov132,LevyG14,LevyGD15}
			\\ \hline		Typical World Knowledge based Approaches    & ESA~\cite{GabrilovichM09}, Probabilistic Conceptualization~\cite{song2011short} & Heterogeneous information networks based meta-path features~\cite{WangChenguangICDM2015} &  OHLDA~\cite{Ha-Thuc11}, KB-LDA model~\cite{Movshovitz-Attias15}, joint text and knowledge embedding~\cite{wang2014knowledge,ZhongZWWC15,xu2014rc,ToutanovaCPPCG15}
			\\ \hline
			Applications                  & Text classification~\cite{GabrilovichM05,GabrilovichM07,GabrilovichM09,Wang14},
			text clustering~\cite{HuJian2008,Hu2009EIE,HuXiaohua2009,Fodeh2011,song2011short},
			information retrieval~\cite{Egozi2011}，
			bag-of-words labeling~\cite{SunXW015},
			search relevance measurement~\cite{Song14},
			search log mining~\cite{Hua2013,WangZWMW15},
			advertising keywords semantic matching~\cite{Liu2012,Kim2013},
			semantic frame identification~\cite{ParkHW16}. & Text classification~\cite{WangChenguangAAAI16}, text clustering~\cite{WangChenguangICDM2015,Wang2015IWK,WangChenguangSRZH16} &  Text classification~\cite{SongR15,LiZTHIS16}, relation classification~\cite{wang2014knowledge}, relation extraction~\cite{ToutanovaCPPCG15}, entity disambiguation~\cite{FangZWCL16}, word analogy~\cite{wang2014knowledge}, recommendation system~\cite{ZhangYLXM16}
			\\
			\bottomrule
		\end{tabular}
	}
	\label{tab:representation}
\end{table*}

\subsection{Explicit Heterogeneous Features}

Instead of treating knowledge base as a source of generating flat features, it is also possible to consider the structural information provided by the knowledge base.
Traditionally, the graph based algorithms only consider the knowledge base as homogeneous graph, and use homogeneous graph based features, e.g., least common ancestor, shortest paths, etc., to disambiguate the words~\cite{BudanitskyH06} and further refine the features in the text documents~\cite{Fodeh2011}.
Even though different kinds of relations can be considered and incorporated, there was no clear framework to formulate them explicitly as graph based features~\cite{Fodeh2011}.
However, when working on the world knowledge graphs, the sparsity of entity relations and computational complexity of finding shortest paths over all possible entities makes shortest path less useful.
In this sense, simpler approaches such as count based features are preferred.
Moreover, traditional approaches focus more on the polysemous and synonymous properties of words, which means focusing more on certain types such as synonym and hyponymy-hypernymy relations.
However, much more types of relations can be considered.
For example, in Freebase, there are thousands of entity types and relations.
A more effective way of using such types and relations should be considered.
In this section, we review the recent development of using heterogeneous information networks to represent the knowledge graph, and using the meta-path to characterize the count-based features through certain relations between entities.
Thus, here we call this approach the ``explicit heterogeneous features.''
It is not a purely graph based feature.
However, developing such features should consider the structure of the graph, as well as a more abstractive level knowledge of the graph.

We first briefly introduce the key concepts related to heterogeneous information network (HIN)~\cite{sun2012mining}. A more comprehensive survey on HIN has been given by~\cite{ShiLZSY17}.

\textit
{
	\begin{definition}
		A \textbf{heterogeneous information network} (HIN) is a graph ${\mathcal G} = ({\mathcal V}, {\mathcal E})$ with an entity type mapping $\phi$: ${\mathcal V} \to \mathcal A$ and a relation type mapping $\psi$: ${\mathcal E} \to \mathcal R$, where ${\mathcal V}$ denotes the entity set and ${\mathcal E}$ denotes the link set, $\mathcal A$ denotes the entity type set and $\mathcal R$ denotes the relation type set, and the number of entity types $|\mathcal A|>1$ or the number of relation types $|\mathcal R|>1$. The \textbf{network schema} for network $G$, denoted as $\mathcal T_{\mathcal G} = (\mathcal A, \mathcal R)$, is a graph with nodes as entity types from $\mathcal A$ and edges as relation types from $\mathcal R$.
	\end{definition}
}

The network schema provides a high-level description of a given heterogeneous information network.
Another important concept, meta-path~\cite{Yizhou11}, is proposed to systematically define relations between entities at the schema level.
\textit
{
	\begin{definition}
		A \textbf{meta-path} $\mathcal P$ is a path defined on the graph of network schema $\mathcal{T}_G = (\mathcal A, \mathcal R)$, and is denoted in the form of ${\it A_1  \xrightarrow{R_1} A_2 \xrightarrow{R_2}  \dots \xrightarrow{R_L} A_{L+1}}$, which defines a composite relation $R = R_1 \cdot R_2 \cdot \ldots \cdot R_L$ between types $A_1$ and $A_{L+1}$, where $\cdot$ denotes relation composition operator, and $L$ is the length of $\mathcal P$. For simplicity, we use type names connected by ``$-$'' to denote the meta-path when there exist no multiple relations between a pair of types: $\mathcal P = {\it (A_1-A_2- \ldots -A_{L+1})}$. We say a path $p = (v_1-v_2- \ldots -v_{L+1})$ between $v_1$ and $v_{L+1}$ in network $\mathcal{G}$ follows the meta-path $\mathcal P$, if $\forall l, \phi(v_l) = A_l$ and each edge
		$e_l = \langle v_l,v_{l+1} \rangle$ belongs to each relation type $R_l$ in $\mathcal P$.
		We call these paths as \textbf{path instances} of $\mathcal P$, denoted as $p \in \mathcal P$.
		\label{def:mp}
	\end{definition}
}

To construct the features based on meta-paths over HIN, the concept of commuting matrix is defined by Y. Sun \etal~\cite{Yizhou11}.
\textit
{
	\begin{definition}
		\textbf{Commuting matrix.}
		Given a network ${\mathcal G} = ({\mathcal V}, {\mathcal E})$ and its network schema $\mathcal T_{\mathcal G}$, a commuting matrix $ {\bf M}_{\mathcal P}$ for a
		meta-path $\mathcal P = {\it (A_1-A_2- \ldots -A_{L+1})}$ is defined as $ {\bf M}_{\mathcal P} = {\bf W}_{A_1A_2} {\bf W}_{A_2A_3}\ldots {\bf W}_{A_{L}A_{L+1}}$, where ${\bf W}_{A_iA_j}$ is the adjacency matrix between types $ A_i$ and $ A_j$. ${\bf M}_{\mathcal P}(i,j)$ represents the number of path instances
		between objects $x_i$ and $y_j$, where $\phi(x_i) = A_1$ and $\phi(y_j) = A_{L+1}$, under meta-path $\mathcal P$.
		\label{def:cm}
	\end{definition}
}

\begin{figure}[t]
	\centering
	\includegraphics[width=0.45\textwidth]{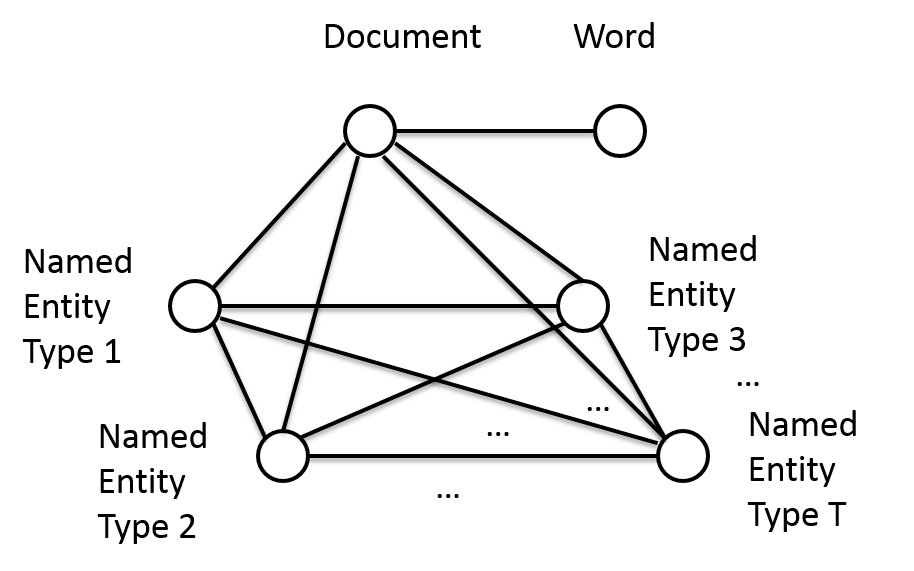}
	\caption{The schema of a document HIN (heterogeneous information network) where the specified knowledge is represented in the form of a heterogeneous information network (Figure from~\cite{Wang2015IWK}). The schema contains multiple entity types: document $\mathcal D$, word $\mathcal W$, named entities $\{{\rm {\mathcal E^I}}\}_{I=1}^T$, and the relation types connecting the entity types.}
	\label{fig:schema}
\end{figure}

For text data, such as a document, we can use semantic parsing and semantic filtering~\cite{Wang2015IWK} to ground the text to world knowledge base.
Then the document can be represented as an HIN.
In addition to the named entities provided by the knowledge base, document and word are also regarded as two types.
Following~\cite{Wang2015IWK}, we use the network schema shown in Fig.~\ref{fig:schema} to represent the data.
The network contains multiple entity types: {\it document} $\mathcal D$, {\it word} $\mathcal W$, {\it named entities} $\{{\rm {\mathcal E^I}}\}_{I=1}^T$, and \emph{relation types} connecting the \emph{entity types}.

By representing the world knowledge in HIN, two documents can be linked together via many meta-paths.
For example, if two documents are linked by the meta-path { {\it Document}$\xrightarrow{\rm{contain}}${\it Politician}$\xrightarrow{\rm{presidentOf}}${\it Country}$\xrightarrow{\rm{presidentOf^{-1}}}$ {\it Politician}$\xrightarrow{\rm{contain^{-1}}}${\it Document}},
the number of the corresponding meta-path instances can be used to measure the similarity between the two documents, which cannot be captured by the original {\it bag-of-words} feature.
We can also represent a document with features defined by meta-paths.
The simplest meta-path is {\it Document}$\xrightarrow{\rm{contain}}${\it Word}.
The calculation based on the meta-paths is to compute all the corresponding commuting matrices of interests.
Then we can derive a lot of general count-based features by the entity types defined in the knowledge graph.
Existing experiments has shown that using this way can help text classification~\cite{WangChenguangAAAI16} and text clustering~\cite{WangChenguangICDM2015,Wang2015IWK,WangChenguangSRZH16} based on very large scale knowledge bases, such as Freebase.

The count based features are very intuitive. However, if the count is not normalized for different meta-paths, it is difficulty to compare over different meta-path based similarities.
Thus, people have thought about the ways of normalization.
One way is called PathSim\etal~\cite{Yizhou11}, where the count based similarity is normalized by the count based similarities of each entity.
Given a symmetric meta path $\mathcal P$, PathSim between two entities $e_x^I$ and $e_y^I$ of the same entity type $I$ is:
$
PathSim(e_x^I,e_y^I) = \frac{2 \times {\bf M}_{\mathcal P}(e_x^I,e_y^I)}
{{\bf M}_{\mathcal P}(e_x^I,e_x^I) + {\bf M}_{\mathcal P}(e_y^I,e_y^I)}.
$
The other way to normalizing the commuting matrix is used in the path ranking algorithm (PRA)~\cite{LaoC10}.
In PRA, instead of normalize the overall commuting matrix, it normalize the individual adjacency matrix.
It assumes that the adjacency defines the transition probabilities between two entities in a certain relation, then the commuting matrix is simulating the commuting time between two entities.
A recent work shows that the random walk defined on meta-paths could be non-stationary, and expected commuting time defined based on random walk is discussed~\cite{HeJiang2017}.

Besides the meta-path, people have also found that meta-graph is very useful when defining the similarities between entities~\cite{FangLZWCL16,HuangZCSML16,HeJiang2017}.
{\it
	\begin{definition}
		A \textbf{meta-graph} ${\mathcal T}_s = ({\mathcal A}_s, {\mathcal R}_s)$ is a sub-graph of network schema $\mathcal{T}_G = (\mathcal A, \mathcal R)$, where ${\mathcal A}_s \subseteq {\mathcal A}$ and ${\mathcal R}_s \subseteq {\mathcal R}$.
		We also denote the subgraph of original HIN as ${\mathcal G}_s = ({\mathcal V}_s, {\mathcal E}_s)$, where ${\mathcal V}_s \subseteq {\mathcal V}$ and ${\mathcal E}_s \subseteq {\mathcal E}$.
		The entities on the subgraph of HIN also follow the mapping $\phi$: ${\mathcal V}_s \to {\mathcal A}_s$ and a relation type mapping $\psi$: ${\mathcal E}_s \to {\mathcal R}_s$.
		\label{def:mp}
	\end{definition}
}
Different from meta-path, where a chain in the network schema is used, meta-graph uses a sub-graph to define the similarities between nodes. However, computing the similarities based on meta-graph is more difficult than meta-path based similarities.

\subsection{Implicit Features}
\label{sec:implicitfeatures}
Analogous to explicit distributional representations, there have been many latent/implicit feature representation for natural language representation.
For example, latent semantic indexing (LSI)~\cite{deerwester1990indexing} has been proposed to work on the explicit features derived by the context.
Later on, probabilistic latent semantic analysis (PLSA)~\cite{Hofmann99} interpret the LSI in a probabilistic way, and further latent Dirichlet allocation (LDA)~\cite{Blei03,Griffiths04} uses a Bayesian model to formulate the generative process of textual documents as bag-of-words.
The key point here is that we can use topic model train the representation of document-topic distributions and topic-word distributions over very large scale, domain independent corpus.
There are many variants of topic models, such as distributed topic models~\cite{NewmanASW09,Surendran06,Porteous08,LiARS14,YuanGHDWZXLM15,0001LZC16}, etc.
Then the topic model can be used to classify the domain dependent documents in the future.

Both PLSA and LDA regard the whole document as word's context, which is more ``global'' consideration compared to other distributional representations~\cite{Hindle1990,Grefenstette92,Grishman1994,Faure98,FaureN99,cimiano2003}.
However, for some natural language processing tasks, such as information extraction and tagging, local contexts are more useful.
To remedy this constrain, neural network language models (NNLMs)~\cite{BengioDVJ03,MorinB05,MnihH08,TurianRaBe10Clean,Collobert11,Mikolov133,Mikolov132,LevyG14,LevyGD15}, or the so called ``distributed word embedding,'' have attracted a lot of attention recently given their compact representation form and generalization property compared to the traditional lexical representations.
Language models~\cite{Manning2008IIR} have been widely used in information retrieval and natural language processing for many years.
However, NNLMs share the advantage of continuous representation of words, and showed the capability of generalization to unseen contexts.

One can argue that in general topic models and NNLMs can incorporate world knowledge if the models are train on the world's available resources.
However, these resources are unstructured, compared to the highly structured knowledge bases/graphs.
In general, when we consider the knowledge base/graph, we want to use the entities and relations, and their types, since they are very semantically useful.
For example, in a knowledge graph, we can find a node ``Microsoft,'' and when we look at its first-hot neighbors, we can retrieve a lot of properties (attributes such as headquarter), related entities (such as its CEO), and similar entities (such as its acquired companies).

Thus, here, we focus on the implicit feature representation that is related to knowledge bases or knowledge graphs.
Compared to explicit features, which are easy to interpret, implicit features are encoded in a way that can not be read by human.
However, implicit features are usually more compact, and has good generalization performance.
In~\cite{Nickel0TG16}, the authors summarized the representation of knowledge graph in the sense of statistical relation learning.
The implicit/latent features of the entities and their relations are mainly used to predict the links of the entities or do inference over the knowledge bases or knowledge graphs themselves.
Note that if we consider link prediction problem being handled by machine learning algorithms, all the approaches surveyed in~\cite{Nickel0TG16} are related, in a generalized way.

Instead of reviewing the knowledge base embedding methodologies, here we emphasize the representation learning algorithms that can generate features based on natural language texts and can be used for other applications.
Such representation can be regarded as incorporating world knowledge since it is not domain dependent, and it can naturally characterize the sparsity and structural information of our knowledge based on the compositional semantics of words.
Moreover, these generated features are more general to be used for machine learning algorithms to work on different tasks.

{\bf Knowledge based topic models.} The most related work is to train a topic model over the knowledge base.
Ontology Guided Hierarchical Latent Dirichlet Allocation (OHLDA)~\cite{Ha-Thuc11} uses class labels in a hierarchy to retrieve documents from Wikipedia, and then trains the topic models on the domain defined by the class labels. In this case, the ontology information are used as queries to submit to search Wikipedia. On the other hand, Wikipedia, as a world knowledge base, severs as an additional source to provide cleaned and relevant documents for the queries. The topic models, in turn, can incorporate the ontology information to guide the topical hierarchy construction, and the topics trained on Wikipedia articles represent the general knowledge about the word distribution of the queries for the topics. The trained topic models can be used for a lot of applications such as text classification~\cite{Ha-Thuc11} as traditional topic models do.
More recently, a KB-LDA model~\cite{Movshovitz-Attias15} was proposed to not only model the hierarchical relations between concepts as OHLDA did, but also model the relations like ``Subject-Verb-Object (SVO)'' to incorporate linked information. It is showed that KB-LDA can better capture richer semantic information in its topics, and show advantage in open IE tasks~\cite{Movshovitz-Attias15}.

{\bf Knowledge enhanced word embeddings.}
Word embedding can also be enhanced by knowledge graphs. For example, joint embedding of words and knowledge graph entities can be performed~\cite{wang2014knowledge,ZhongZWWC15,xu2014rc,ToutanovaCPPCG15,AhnCPB16} by first aligning text with knowledge graph (which may be imperfect), combining the objective functions of word embedding and knowledge graph embedding, and jointly optimize both together.
Moreover, the similar approaches can be used to improve knowledge graph embedding~\cite{wang2014knowledge,FangZWCL16}.
These approaches are interesting since they are related to the new learning paradigms we will introduce later in Section~\ref{sec:paradigms}.
When using the knowledge graph to improve the word embedding, it is highly related to distant supervision and indirect supervision.
For distant supervision, it means the supervision of entity embedding from knowledge graph is incorporated in an inexact way. The alignment of entities and unstructured texts is not perfect. Thus, different entity senses can bring noise in the word embedding results.
For indirect supervision, it means the supervision of entity embedding is not directly used to supervise words. Instead, the relation embeddings are shared with word embedding and knowledge graph embedding objective functions, while the entities shown in the text are freely optimized based on composition of word embeddings.

{\bf Combined representations.}
One can also combine the explicit and implicit representations of knowledge graph to improve the results.
For example, the ESA can be simply augmented by considering a bag-of-concept-embeddings representation~\cite{SongR15}.
This approach is later refined by directly incorporate the knowledge graph embedding into ESA~\cite{LiZTHIS16}.
It has been shown that this approach can be more robust than original ESA, especially when the number of concepts used in ESA is chosen to be small.

A summary of traditional text representation and knowledge enhanced models and applications are shown in Table~\ref{tab:representation}.

\section{Inference with World Knowledge}
\label{sec:inference}
In this section, we review the inference techniques related to world knowledge.
To incorporate a world knowledge base in either representation or learning, it is important as a first step to link the free texts to the knowledge base entities and relations (which are also called grounding).
We call these tasks inference because when assigning labels to the entities or phrases we need to look at global information in a document or even in a corpus.
This cannot be simply learned but to be inferred based on statistics and constraints.
When doing inference, there are two most important issues need to be considered.

{\bf Ambiguity.} Similar to the polysemy of words, the entities or relation expressions in free texts can express multiple meanings. For example, ``Alex Smith''  can refer to ``Quarterback of the Kansas City Chief'' or ``Tight End of the Cincinnati Bengals.''
It will be the context to determine the real reference when the entities are mentioned.

{\bf Variability.} The other problem is that a given concept can be expressed in many ways, which is similar to the synonyms of words.
For example, ``cat'' can be expressed by ``feline,'' ``kitty,'' and ``moggy.''

When we consider grounding the free text to the world knowledge bases, we need to carefully consider the above two problems.
In this section, we review two important problems related to inference with world knowledge, i.e., entity linking and semantic parsing.

\subsection{Entity Linking}
A comprehensive survey of different approaches of entity linking has been given by~\cite{ShenWH15}.
Here we focus on the inference problem in entity linking to discuss the entity linking in two perspectives: local and global inference.

We first introduce the notations and definition of entity linking.
We first define a mention (a concept or an entity) detected or needed to be highlighted in free text as $e$.
The mentions in free text can be multi-word expression, referring to named entities (person, location, organization), objects, events, philosophy, mental states, rules, etc.
Then we determine what is the target encyclopedic resource (or knowledge base), e.g., Wikipedia or Freebase.
Then the task of entity linking is to define what mentions to point to the entities/concepts in the knowledge base.
More specifically, we define the title of an entity or a concept in the knowledge base as $c_i\in \CM =\{c_1, \ldots, c_T\}$.
Here we use $c_i$ to be consistent with Section~\ref{sec:representation} where the title can refer to either a concept or an entity.
For example, in Example 3, we can detect ``Martha Stewart'' linked to Wikipedia, which is categorized as ``person,'' ``founder,'' ``winner,'' etc.
When the mentions are linked to Wikipedia, the task is also called Wikification.
There is also a subtle difference between entity linking and Wikification.
When mentions cannot be linked to Wikipidia, Wikification only returns ``Null'' as the link, while entity linking task also requires the program to cluster the relevant mentions to represent a unique concept and map the cluster to certain ``Null'' category.
Compared with ESA introduced in Section~\ref{sec:representation} which links to many related concepts as a text representation, Wikification only links to the best candidate of concept.
In this section, we do not discuss the difference between the tasks but only focus the inference problem involved.

\subsubsection{Mention Identification}
Before linking to the knowledge bases, the first step is mention detection or identification.
This is a non-trivial task since the our natural language is arbitrary, and the boundary of mentions is also arbitrary.
Thus, a lot of approaches have been proposed in different ways, e.g., using shallow parsing to find NP (noun phrase) chunks, leveraging the named taggers~\cite{RatinovR09}, developing specific mention extractors~\cite{FlorianJKZ06,LiJ14}, considering only n-grams~\cite{RatinovRDA11}, and a lot of other methods~\cite{RothJCC14}.
Existing systems include Illinois Wikifier~\cite{RatinovR09,ChengR13}, which uses NP chunks and sub-strings as candidates, and uses prior anchor texts to determine other potential string; TAGME~\cite{FerraginaS10}, which uses prior anchor texts to identify mentions; DBPedia Spotlight~\cite{MendesJGB11}, which uses dictionary based chunking with string matching to DBPedia lexicon; AIDA~\cite{YosefHBSW11}, which uses name tagging system for mention detection; and RPI Wikifier~\cite{ChenJ11,CassidyJRZH12,CassidyJDZH12,HuangCHJL14}, which uses mention extraction sub-routine to detect mentions~\cite{LiJ14}.
A comparison of different approaches is presented in~\cite{MendesDRSB12}.

\subsubsection{Local Inference}
Given a set of mentions being detected, the entity linking task mainly considers to link the mentions to the entities or concepts in the knowledge base (for Wikipedia, the titles).
Thus, in general, if we have a mention $e_m$, and a set of entities or concepts $c_t\in \CM $,
then local inference uses the mention itself and a set of local context features to determine which entity or concept it refers to.
For example, we can use the joint probability $P(e_m, c_t)$, and conditional probabilities $P(c_t | e_m)$ and $P( e_m | c_t)$ to rank the candidates.
The probability $P(c_t | e_m)$ characterizes the ``commonness'' of a mention referring to a title~\cite{MedelyanMLW09}.
If we see a mention ``Chicago'' in the text, the probability of $P(``title" | Chicago)$ is used to rank the titles such as ``Chicago'' as a city, ``Chicago (band),'' ''Chicago (2002\_film),'' etc.
This is very related to the method mentioned in probabilistic conceptualization in Section~\ref{sec:representation}.
While PC uses $P(c_t | e_m )$ to generate a lot of concepts to describe the entity, here this probability is used to rank the best one for the mention.
This method is usually used as an initial ranking since it is not robust across different domains.
For example, in ~\cite{RatinovRDA11,MeijWR12}, the results have shown that for different topics/genres and different domains (e.g., news and tweets), the performance diverse a lot.
Some extension of initial count based ranking using graph based features were also proposed~\cite{HacheyRC11,HakimovOD12}.

To further improve the local inference results, more complicated contextual features have been proposed.
In general, the features can be used to compute the similarity between the mention and the title $\phi (e_m, c_t)$, then the overall inference is to solve the following maximization problem:
\begin{equation}
f^*_{m\rightarrow t} = \arg \max_{f_{m\rightarrow t}} \sum_{e_m \in d} \phi (e_m, c_t).
\end{equation}

There are multiple ways to define the similarity function $\phi (e_m, c_t)$ based on features, including
the name string matching, document surface, entity context, concept, KB link features, profiling, topic, popularity, etc.~\cite{ZhengLHZ10,DredzeMRGF10,AnastacioMC11,ChenTLLLSAPJ10,ChenJ11,CassidyJDZH12,MilneW08,SyedFJ08,SrinivasanCS09,Kozareva2011,ZhangSST11,AnastacioMC11,CassidyCAJDRHRZ11,PinkNRCNTC13}.
In general, if the features have been constructed, some traditional similarity metrics such as cosine or Jaccard can be used.
There are also other approaches such as mutual information or second order vector composition used for similarities evaluation~\cite{HoffartYBFPSTTW11}.
Given the above similarities, we can further run learning to rank~\cite{RatinovRDA11,ChenJ11} to find the weights for each feature to rank the candidates of titles for the mentions.
There are a lot of different supervised or weakly supervised learning approaches~\cite{JiG11,RothJCC14,ShenWH15}.
Here machine learning is used as a tool for local inference.
The key idea here is that the concepts are evaluated based on the features extracted from the context to link the mentions to the knowledge base.
Again, compared to ESA and probabilistic conceptualization introduced in Section~\ref{sec:representation}, here much more features are used to disambiguate the concepts.
In ESA or probabilistic conceptualization only simple string matching~\cite{GabrilovichM09} or very shallow parsing~\cite{Song14} are considered.

\subsubsection{Global Inference}\label{sec:linking-global}
Besides directly comparing the entity mentions in a document and the candidate concepts in the knowledge base, the concepts in a document can also help disambiguate each other.
For example, ``Chicago VIII'' (being an album) can be used to disambiguate the entity ``Chicago'' (being a band) when they are mentioned together in a document.
Thus, a framework of global inference needs to be developed to jointly optimize the entity mention and concept candidate similarities as well as the concepts relatedness in a document:
\begin{equation}
f^*_{m\rightarrow t} = \arg \max_{f_{m\rightarrow t}} \sum_{e_m \in d} [\phi (e_m, c_t) + \sum_{c_t, c_{t'} \in d}\varphi (c_t, c_{t'})].
\end{equation}
By comparing with local inference, a new term $\varphi (c_t, c_{t'})$ is added to each entity mention $e_m$ in the document.
For a candidate $c_t$, all the possible related concepts $c_{t'}$ shown in the document $d$ is evaluated.
If more concepts in the documents are related to $c_t$, then it is more likely to be the concepts the entity mentioning.
The first relatedness score between concepts was developed in~\cite{MilneW08}.
Then it is used by many systems~\cite{KulkarniSRC09,RatinovRDA11,HoffartYBFPSTTW11}.
More relatedness scores are also developed.
A comprehensive study of different scores is given by~\cite{CeccarelliLOPT13}.

\subsection{Semantic Parsing}
Entity linking only works on linking the entity mentions in free texts to the knowledge base.
However, the relations between the entities are not considered.
If we also want to map their relations in the text to the knowledge base, semantic parsing should be developed.
Traditionally, semantic parsing refers to the task of mapping a piece of natural language text to a formal meaning representation~\cite{Mooney07}.
In different context, semantic parsing can mean different tasks.
For example, when there is no knowledge base to be grounded, semantic parsing can be in the form of CCG (Combinatory Categorial Grammar) parsing~\cite{ArtziFZ13} or shallow semantic role labeling~\cite{2010Palmer}.
Formally, we convert a given sentence to the most appropriate logic forms.
For example, for the two questions shown below, we can convert them to different logic forms:

{\bf Example 4~\cite{KwiatkowskiCAZ13}: natural language texts and their logic forms.}

{\it Example 4.1: What is the population of Seattle?}

{\it Example 4.2: How many people live in Seattle?}

{\it Example 4.3: $\lambda x. population(Seattle, x)$}

{\it Example 4.4: $count(\lambda x.person(x)\wedge live(x.Seattle))$}

Both Examples 4.1 and 4.2 refer to the same meaning. However, their logic forms could be different (Examples 4.3 and 4.4), depending on which semantic parsing algorithm we rely on, as well as the lexicon we can build to determine the paraphrasing similarities between predicates~\cite{KwiatkowskiCAZ13}.

However, such semantic parsing does not provide the fine-grained entity types as knowledge bases do.
When grounding to knowledge bases, semantic parsing is well known for question answering.
Most previous semantic parsing algorithms or tools developed are for small scale problems but with complicated logical forms~\cite{KwiatkowskiZGS11}.
More recently, large scale semantic parsing grounding to world knowledge bases has been investigated, e.g., using Freebase~\cite{Krishnamurthy2012,CaiY13,KwiatkowskiCAZ13,berant2013semantic,BerantL14,YaoD14,ReddyLS14} or ReVerb~\cite{FaderZE13}.
More formally, let $\mathcal{E}$ be a set of entities and $\mathcal{R}$ be a set of relations in the knowledge base.
Then a knowledge graph $\mathcal K$ consists of triplets in the form of $(e_1, r, e_2)$, where $e_1, e_2 \in \mathcal E$ and $r \in \mathcal R$.
Here we take Lambda Dependency-Based Compositional Semantics~\cite{berant2013semantic} as an example to demonstrate the specific forms of grounding natural language texts to logic forms with types.

{\bf Example 5: Lambda Dependency-Based Compositional Semantics~\cite{berant2013semantic} ($\lambda$-DCS)~\cite{Liang13}.}

The simplified L$\lambda$-DCS~\cite{Liang13} defines the logic language to query the knowledge base.
The logical form in simple $\lambda$-DCS is either in the form of unary (a subset of $\mathcal E$) or binary (a subset of $\mathcal E \times \mathcal E$).
We briefly introduce the definition of basic $\lambda$-DCS logical forms $x$ and the corresponding denotations $x_{\mathcal K}$ as below:
(1) Unary base: an entity $e \in \mathcal E$ is a unary logic form (\eg, {\it Seattle}, {\it University}) with $e_{\mathcal K} = \{e\}$; (2) Binary base: a relation $r \in \mathcal R$ is a binary logic form (\eg, {\it Type}, {\it Education}, {\it PlaceOfBirth}) with $r_{\mathcal K} = \{(e_1,e_2):(e_1,r,e_2)\in \mathcal K\}$; (3) Join: $b.u$ is a unary logic form, denoting a join and projection, where $b$ is a binary and $e$ is a unary. ${b.u}_{\mathcal K} = \{e_1 \in \mathcal E : \exists e_2.(e_1,e_2) \in b_{\mathcal K} \wedge e_2\in u_{\mathcal K}\}$ (\eg, {\it Type.University}, {\it Education.BarackObama}, {\it PlaceOfBirth.Obama}); (4) Intersection: $u_1 \sqcap u_2$ ($u_1$ and $u_2$ are both unaries) denotes set intersection: ${u_1 \sqcap u_2}_{\mathcal K} = {u_1}_{\mathcal K} \cap {u_2}_{\mathcal K}$ (\eg, {\it Type.University $\sqcap$ Education.BarackObama}).
\nop{For example, ``Gone with the Wind is written by Margaret Mitchell'' can be represented as {\it {Book.Written$\wedge$Work.People.Person}} in $\lambda$-DCS. {\color{red} this is not in the correct form! we need a better example here.}}
Then overall, for the text below, we can parse the logic forms
Most grounding based semantic parsing approaches are applied to answer questions with the world knowledge bases~\cite{berant2013semantic,BerantL14,YaoD14,ReddyLS14}. For example:

{\it Example 5.1: Who is the president of United States.}

{\it Example 5.2: $Type.People\sqcap PresidentofCountry.USA$.}
Here the two relations $PresidentofCountry$ and $Country.USA$ join together to have $PresidentofCountry.USA$.
Moreover, the word ``Who'' is lexically mapped to $Type.People$ where $Type$ is the predicate of the ``isA'' relationship of entity and its concept.
Thus, both the relationship between entities and the type information (or higher level concepts) of entities are naturally incorporated into the logic form by grounding to the knowledge graph.

Similar to entity linking, semantic parsing also has different approaches, i.e., local inference based and global inference based approaches.

\subsubsection{Local Inference}
Most of the existing semantic parsing approaches are local inference based, since most of them are applied to question answering, which only targets to parse very simple sentences.
There are multiple ways of finding a proper semantic parsing results for natural language texts.
First, local inference involves mapping the entity mentions and relation expressions in natural language texts to the entities and relations (or predicates) in the knowledge base.
The entity identification is similar to the mention identification and linking problems in entity linking.
However, for knowledge bases like Freebase, Yago, there are less texts to describe the entities as Wikipedia has.
Therefore, the entity identification and linking problems are more difficult.
For the relation expressions, paraphrasing~ are usually used to identify the relations~\cite{FaderZE13,BerantL14}.
To identify the correct logic forms from text, the typical way is to determine a logic form based on some cost function:
\begin{equation}
z^* = \arg\min_{z \in {\mathcal K}} \psi (z, s, w^*),
\end{equation}
where $z$ is a possible logic form that can be mapped to a sub-graph of the knowledge graph, $s$ is the give natural language sentence (or utterance), and $w$ is the parameter in the parameter space $\Theta$ to minimize the cost function:
\begin{equation}
w^* = \arg\min \sum_{z \in {\mathcal K}} \ \psi (y, z, s, w^*),
\end{equation}
where $y$ is possible supervision for the logic forms.
Different from entity linking, there is no naive unsupervised similarity to evaluate the candidate logic form and the sentence.
Therefore, to find a proper logic form for the give text, different strategies has been investigated.

The most intuitive way is to use supervised learning to learn the parameters for the cost function, when the correct logic forms are annotated for the sentence~\cite{WongM07,KwiatkowksiZGS10,YihRMCS16}.
However, such supervision heavily requires laboring cost since the grounded logic forms in the knowledge graph can be exponentially many for a human to judge.
Thus, a lot of studies have been done to reduce or replace the annotation requirement for such problems.

Instead of direct supervision of logic forms, indirect supervision from the answer can be used to train the parameters~\cite{ClarkeGCR10,LiangJK11}, where the logic forms are treated as hidden variables in the cost function.
Then to optimize the cost function, latent variables should be integrated out when working with parameters.
There could be exponentially increasing logic forms candidates extracted from the sentence.
Thus, heuristic pruning or beam search should be performed~\cite{berant2013semantic,BerantL14}.
Recently, staged parsing has been investigated~\cite{YihCHG15}, which reports more efficient and effective results.
Other learning strategies such as combining imitation learning and agenda based parsing can also be used to improve the efficiency and effectiveness of semantic parsing, which also significantly reduce the search space~\cite{BerantL15}.

Distant supervision (which will be introduced in Section~\ref{sec:paradigms}) can also be used to supervise semantic parsing.
Here distant supervision refers to the approach using the knowledge base to find entity relation triples $(e_1, r, e_2)$ in Web scale documents, and then use high frequent triples as ``gold'' to supervise the lexical mapping from knowledge graph entities and predicates and the texts where triples exists~\cite{Krishnamurthy2012,CaiY13,ReddyLS14}.
For example, Google released a data set using Freebase to automatically annotate the large Web document collection ClueWeb~\cite{Gabrilovich13FreebaseClueWeb}.
In this way, we can obtain a lot of cheap annotation, but the annotation is very noisy and incomplete when training the model.
Some neural network based learning models has been proposed to replace some components or the final learning algorithm of grounded semantic parsing~\cite{YihCHG15}.

Some systems or approaches also consider an ``unsupervised'' way to train the parameters~\cite{GoldwasserRCR11,FaderZE13,Poon13}.
To our best knowledge, \cite{GoldwasserRCR11} was the first grounded unsupervised semantic parsing, which adopted a ``self-training'' strategy.
It uses some initial seeds which is evaluated by a translation based score (which also relies on a lexicon related to the knowledge base predicates), and further bootstraps the learning procedure to update the system parameters.
Fader et al.~\cite{FaderZE13} use a paraphrasing to evaluate similarities between questions and the possible grounded results, where the grounding should be restricted to simple forms, such as unary and binary relations.
Poon~\cite{Poon13} uses dependency tree as the backbones of candidate logic forms, and tries to annotate the tree with knowledge base entities and relations (predicates).
All the above approaches assume that given a set of questions, the logic forms can be induced by maximizing the likelihood of certain constrained logic forms.
Thus, they are still limited to either small scale knowledge graphs or simple logic forms.

\subsubsection{Global Inference}

When semantic parsing are applied to a document, global inference can also be performed.
This is very similar to the unsupervised approaches introduced in local inference.
However, local inference leverages a set of question to determine the parameters, but not uses the relationships among entities and relations extracted from the questions to disambiguate each other.
As entity linking, the logic forms can be filtered when we know a lot of logic forms from other sentences in a document or a corpus.

Here we show a semantic filtering strategy proposed in~\cite{Wang2015IWK,WangChenguangSRZH16}.
For example, motivated by the approaches of ``generative+discriminative'' conceptualization~\cite{song2015short}, we can represent each entity with a feature vector ${\bf c}_i = (c_1, \ldots, c_{{T}})^T$ of entity types and use standard Kmeans to cluster the entities in a document. Suppose in one cluster we have a set of entities $ E = \{e_1, \cdots, e_{N_{E}}\}$. Then we can use the probabilistic conceptualization proposed in~\cite{song2011short} to find the most likely entity types for the entities in the cluster. We make the naive Bayes assumption and use $ P(c_k|\mathcal E)  \propto P(c_k)\prod_{i=1}^{N_{\mathcal E}}P(e_i|c_k)
$
as the score of entity type $c_k$. Here, $P(e_i|c_k) = \frac{n(e_i,c_k)}{n(c_k)}$ where $n(e_i,c_k)$ is the co-occurrence count of entity type $c_k$ and entity $e_i$ in the knowledge base, and $n(c_k)$ is the overall number of entities with type $c_k$ in the knowledge base. Besides, $P(c_k) = \frac{n(c_k)}{\sum_k^{{T}} n(c_k)}$.
The probability $P(c_k| E)$ is used to rank the entity types and the largest ones are selected.
In this case, for each cluster of entities, only the common types are retained, and concepts with conflicts are filtered out.
It is also possible to apply the global inference approaches used in entity linking shown in Section~\ref{sec:linking-global}.

\section{Learning: Related Learning Paradigms}
\label{sec:paradigms}
In this section, we introduce the new learning paradigms that are enabled by world knowledge.
We categorize the paradigms into ways related to world knowledge features and ways related to world knowledge supervision.
For the paradigms related to world knowledge features, the representations introduced in Section~\ref{sec:representation} are incorporated into the learning algorithms.
For the paradigms related to world knowledge supervision, the inferences are mainly used to find the categorized entities and relations that are inferred by the approaches introduced in Section~\ref{sec:inference}.

\begin{table*}[htbp]
	\centering
	\caption{Comparison of learning paradigms.}
	\begin{tabular}{c|p{0.12\textwidth}|p{0.12\textwidth}|p{0.12\textwidth}|p{0.12\textwidth}}
		\toprule
		& Labeled data in training & Unlabeled data in training & Label names in training &  Training and testing I.I.D. \\
		\midrule
		Supervised learning                     & $\surd$ &  &  & $\surd$ \\
		\hline
		Unsupervised learning                   &  & $\surd$ &  & $\surd$ \\
		\hline
		Semi-supervised learning                & $\surd$ & $\surd$ &  & $\surd$ \\
		\hline
		Self-taught learning	                & $\surd$ & $\surd$ &  &  \\
		\hline
		Traditional transfer learning                       & $\surd$ (no for unsupervised transfer) & $\surd$ &  &\\
		\hline
		Source free transfer learning           &  $\surd$ (optional)  & $\surd$  &  $\surd$ &  \\
		\hline
		One-shot learning                       & 1 instance per class &  &  & $\surd$ \\
		\hline
		Zero-shot learning                      & $\surd$ &  $\surd$ & $\surd$ &\\
		\hline
		Dataless classification                 &  &  & $\surd$ &\\
		\hline
		Dataless classification (Bootstrapping) &  & $\surd$ & $\surd$ & $\surd$ \\
		\midrule
		Distant supervision                    &  $\surd$  &  &   & $\surd$ \\
		\hline
		Indirect supervision                    &  $\surd$  &  &   & $\surd$ \\
		\bottomrule
	\end{tabular}
	\label{tab:learningparadigm}
\end{table*}

\subsection{Paradigms Enabled by World Knowledge Features}

We first review the new learning paradigms that are enabled by the features generated by world knowledge.

\subsubsection{Self-taught Learning}

The first learning setting enabled by universal world knowledge is called ``self-taught learning''~\cite{RainaBLPN07,LeeRTN09}.
Self-taught learning uses a large amount of unlabeled data crawled from the Web to train an unsupervised representation learning.
The a supervised classifier can be applied to the features trained based on the unlabeled data.
It can be regarded as using a lot of data to find a better universal data distribution $P(\mathcal{X})$, which may not be strongly related to $P(\mathcal{Y}|\mathcal{X})$.
Then the discriminative classifier is fine-tuned for $P(\mathcal{Y}|\mathcal{X})$.
Essentially this is a semi-supervised learning setting, where large amount of unlabeled data is used to help supervised learning tasks with less labels.
Particularly, self-taught learning decouples the representation learning part using unlabeled data and classifier training using labeled data, and it does not require that the labeled data and unlabeled data are sampled from the same distribution.
ESA applied to text classification also shares this ides~\cite{GabrilovichM05,GabrilovichM06,GabrilovichM07,GabrilovichM09}, where the features for a piece of short text can be generated from Wikipedia, and then the classification is performed over the knowledge based features.

The deep learning community also found unsupervised learning using restricted Bolzman machines (RBM)~\cite{HintonOT06}, auto-encoders~\cite{BengioLPL06,RanzatoPCL06,LarochelleECBB07}, and their variants (see~\cite{Bengio09} for more comprehensive survey) are helpful for training very deep neural network architectures.
They use unsupervised learning as pre-training trained on a lot of unlabeled data, and then introduce a fine-tuning process to refine the model with very deep architectures.
In natural language processing, it has also been shown that using the Brown clusters of words~\cite{PLiang05} or word embeddings~\cite{TurianRaBe10Clean,Collobert11} learned from large scale of unlabeled data, and training another classifier (either traditional~\cite{PLiang05} or deep learning~\cite{TurianRaBe10Clean,Collobert11}) for specific tasks can help improve the task specific performance.
Thus, the representations introduced in Section~\ref{sec:implicitfeatures} can be regarded as a pre-training step or a self-taught learning step for many NLP applications.

\subsubsection{Source-free Transfer Learning}
Originally, as shown in Section~\ref{sec:transferlearning}, transfer learning refers to the setting of training with a source domain containing a lot of labeled data.
However, there are two major challenges.
One is the availability of source domain data, and the other is how to automatically select a source domain for transfer learning.
Source-free transfer learning tries to solve this problem with world knowledge instead of domain knowledge~\cite{XiangPPSY11,LuZPXW014}.

One way to use world knowledge in source-free transfer learning is to use the Wikipedia categories~\cite{XiangPPSY11}.
Wikipedia categories provide large amount of categorized text data.
Then a large set of classifiers can be built based on the categorized data.
For a new coming target domain (in text, other format of data needs do consider heterogeneous transfer learning~\cite{Pan2010}), the label similarities can be evaluated between source domains and target domain to automatically find a source domain to transfer.
The other way of using world knowledge is that when an initial classifier can be built for a target domain, then a meta-learning algorithm can be designed to automatically use the key features (keywords) in the classifier to query more unlabeled data from cleaned Wikipedia corpus~\cite{LuZPXW014}.
Then traditional semi-supervised learning algorithms, such as graph regularization~\cite{BelkinNS06}, can be applied to iteratively train a new classifier based on the labeled and incrementally increasing unlabeled data.
OHLDA~\cite{Ha-Thuc11} can also be regarded as in the source-free transfer learning framework.
It uses the topical label keywords as search query to search Wikipedia or Google, and then uses the retrieved documents to train a hierarchical topic model.
The topic model can be used as the classifier to classify the documents to the given ontology of labels.

\subsubsection{Dataless Classification with Semantic Supervision}

Dataless classification performs a nearest neighbor search of labels for a document in an appropriately selected semantic space~\cite{ChangLRS2008,SongR14}.
Let $\phi(d)$ be the representation of document $d$ in a semantic space (to be defined later) and let $\{\ph(l^{(1)}),\ldots,\ph(l^{(N_l)})\}$ be the representations of the $N_l$ labels in the same space.
Then we can evaluate similarity using an appropriate metric $f(\ph(d), \ph(l^{(i)}))$, (e.g.,
cosine similarity between two sparse vectors) and select label(s) that maximizes the similarity: $l^{*} = \arg \max_i f(\ph(d) ,
\ph(l^{(i)}))$.
Essentially this learning paradigm should be called ``supervisionless'' or semantic supervision with label names.

The core problem in dataless classification is to find a semantic
space that enables good representations of documents and labels.
Traditional text classification makes use of a bag-of-words (BOW)
representation of documents. However, when comparing labels and
documents in dataless classification, the brevity of labels {makes} this simple minded representation and the resulting similarity measure unreliable. For example, a
document talking about ``sports'' does not necessarily contain the
word ``sports.''
Consequently, other more expressive distributional representations have been applied, e.g., Brown
cluster~\cite{BrownPdLM92,PLiang05}, neural network
embedding~\cite{Collobert11,TurianRaBe10Clean,MikolovYZ13,Mikolov132},
topic modeling~\cite{Blei03}, ESA~\cite{GabrilovichM09}, and their
combinations~\cite{SongR15}.  It has been shown that ESA gives the best
and most robust results for dataless classification for English
documents~\cite{SongR14}.
This idea can be generalized to hundreds of languages with Wikipedia with language links to English~\cite{SUPR16} where both English labels and documents in other languages can be mapped to the same semantic space using cross-lingual ESA~\cite{Potthast2008,Sorg2012}.
Then it can be further extended to classify any languages in the world with a dictionary mapping from the target language and a pivot language that can be linked to English~\cite{SongDataless2017}.

Zero-shot learning ~\cite{PalatucciPHM09,SocherGMN13,Elhoseiny13,RomeraParedesT15} were also introduced in the computer vision community and are now recognized by the natural language processing community~\cite{YazdaniH15,LazaridouDB15,JohnsonSLKWCTVW16}.
Zero-shot learning means that there is no labeled data in the new coming target domain.
However, it requires some source data to train the model.
The target classifier is bridged based on the label similarities between target labels and the source labels.
In this sense, zero-shot learning is very similar to the first mechanism of source-free transfer learning.
Another learning mechanism with similar name is one-shot learning~\cite{FeiFeiFP06,Brenden2015}.
However, one-shot learning requires one example for training, while in zero-shot learning, the test data is different from the training data (e.g., a new label space).

A comparison of different related learning paradigms is shown in Table~\ref{tab:learningparadigm}.

\subsection{Paradigms Enabled by World Knowledge Supervision}

World knowledge can not only be used as features, but also used as supervision. In this section, we review two learning paradigms that are enabled by world knowledge supervision, i.e., distant supervision and indirect supervision.

\subsubsection{Distant Supervision}

The idea of using minimal supervision not aligned to the entity mentions in texts (or weak supervision from domain knowledge bases)~\cite{BunescuM07} has been explored previously.
Distant supervision extends this idea to use the knowledge of entities and their relationships from world knowledge bases, e.g., Freebase, as supervision for the task of entity and relation extraction~\cite{mintz2009distant}.
Since the entities and their relations in the world knowledge bases are not aligned with the mentions in the texts, the first step of distant supervision is to find the entities in the texts.
Entity linking can be considered, but cheaper ways such as simple string matching or named entity tagger~\cite{mintz2009distant} can be employed since for the training step, only the most confident examples are needed.
Then the sentences with mapped entities and their relations can be used as labeled examples to train a relation classifier.
Since there is no direct annotation about the entities and relations in the sentence but only the automatically mapped annotation from knowledge base is used, this approach is called distant supervision.
During the whole process, one can argue that no human annotation (directly related to the task) is needed for the (open-domain) relation extraction problem.

The automatically aligned entities are not accurate enough for relation extraction.
If we assume that for multiple sentences with the same pair of entities extracted, only part of them support the decision of relation between the two entities, then the problem can be formulated as a multi-instance learning problem~\cite{RiedelYM10,XuHZG13}.
It is proved that multi-instance learning can significantly reduce the effect of noisy labels.
Moreover, since two entities can have more than one type of relations, the distance supervision can also be formulated as multi-instance multi-label learning problem~\cite{HoffmannZLZW11,SurdeanuTNM12}.

Another view of relation extraction with knowledge distant supervision is to formulate the problem as a matrix completion problem~\cite{RiedelYMM13,FanZZLZC14}, which is also related to other statistical representation learning with knowledge graph studies~\cite{Nickel0TG16}.
This approach claims they can unify the open information extraction (openIE) and relation classification.
Moreover, it has been shown that the joint representation of words and knowledge graph can also help improve the distant supervision for relation classification~\cite{wang2014knowledge}.
More recently, neural network learning based algorithm has been tested on distant supervision setting~\cite{LiuLXZ14,ZengLC015,LinSLLS16}.

Other extensions of different learning settings have also been proposed.
For example, the combination of semi-supervised learning~\cite{AngeliTWM14,PershinaMXG14} and background knowledge~\cite{NageshHR14,RocktaschelSR15} can also help distant supervision's performance.
\cite{MinGW0G13} also uses positive and unlabeled learning (PU-learning) setting to handle the distant supervision when the knowledge base is incomplete.

\subsubsection{Indirect Supervision}
Besides the distant supervision, it is also possible to use world knowledge as indirect supervision.
The idea of indirect supervision is a general learning setting~\cite{ChangSGR10,LiangJK11,RaghunathanFDL16}.
For example, \cite{ChangSGR10} uses the cheaper annotation as indirect supervision for more complicated learning problems.
The document-level topic annotation is cheaper than the information extraction (named entities, events, etc.) annotation.
Then the document-level binary or multi-class classification can help refine the parameter esitmation problem of structural output learning.
\cite{LiangJK11} uses answers to supervise the semantic parsing problem in question answering.
The target of semantic parsing is to output the formal logic forms.
However, it is too costly to label a lot of logic forms for machines to learn.
Thus, using indirect supervision is a natural way to reduce the labeling work.
\cite{RaghunathanFDL16} further introduced a new learning setting for privacy preserved machine learning based on indirect supervision.
In the context of natural language processing, as an example, they use labeling the number of particular POS tags (such as noun) instead of labeling individual tags as indirect supervision.
In summary, different from transfer learning, indirect supervision can be more different when comparing the available annotation and the target of machine learning.
It is not a general learning setting but should be introduced case by case.

In the research related world knowledge, Song et al.~\cite{song2013constrained} considered using fully unsupervised method to generate constraints of words using an external general-purpose knowledge base, WordNet, for document clustering.
This can be regarded as an initial attempt to use general knowledge as indirect supervision to help clustering.
However, the knowledge from WordNet is mostly linguistically related.
It lacks of the information about named entities and their types.
Moreover, their approach is still a simple application of constrained co-clustering, where it misses the rich structural information in the knowledge base.
To extend this idea, indirect supervision using world knowledge bases such as Freebase and Yago has been proposed and applied to document clustering problem~\cite{Wang2015IWK}.
It uses semantic parsing to ground the entities and their relations to the world knowledge base, and builds a heterogeneous information network to formulate the structured data for the documents.
Then the information about the entity categories, e.g., celebrities, IT companies, politicians, can be propagated to group the documents with topics.

\section{Conclusion and Future Directions}

In this paper, we have formulated the problems of machine learning with world knowledge, and reviewed the related methodologies and algorithms involved in the respective problems.
We first compare learning with world knowledge to learning with domain knowledge, and then we categorize the problems into three folds, i.e., representation, inference, and learning.
For representation, we introduced different kinds of representation with world knowledge, which are explicit homogeneous features, explicit heterogeneous features, and implicit features.
For inference, we introduced entity linking and semantic parsing to align free text to knowledge bases.
For learning, we introduced several new learning paradigms that are enabled by world knowledge.
There are still a lot of open problems that have no answer at the current stage, which we think are very important, including:
\begin{itemize}
	\item {\bf Commonsense acquisition and learning with commonsense knowledge}. Commonsense knowledge has been a key problem since the artificial intelligence has been introduced. World knowledge can cover partial commonsense knowledge but there is still a lot of such knowledge missing. Commonsense knowledge is very important when performing inference about natural language as well as many other applications~\cite{Davis2015}. Thus, how to acquire and use commonsense knowledge are still open problems.
	\item {\bf Representation or representation learning}. Currently there has been a lot of comparison between traditional distributional representation and advanced representation learning. For some of the tasks, such as dataless classification, traditional distributional representation, i.e., ESA, still out performs learning based representations. It would be very important to find out ways to further improve the corresponding representation using the more advances learning techniques.
	\item {\bf Joint inference and learning}. The problems inference and learning introduced in this paper are mostly separated. Joint inference and learning may help each other to boost the performance~\cite{RocktaschelSR15}. We regard this as a natural idea of a general machine learning with world knowledge framework. The challenge is that the joint inference and learning will be very costly both in terms of computational efficiency and effectiveness.
	\item {\bf Cross-lingual and cross-culture world knowledge}. Different communities of people with different background may have different kind of common knowledge~\cite{SUPR16}. The information collected from the Web about the world knowledge can be biased and different for different languages and cultures. It will be interesting to compare different languages and cultures, and find ways to correct the bias for them.
	\item {\bf Connecting world knowledge to cognitive science}. There are evidence about human performing transfer learning~\cite{Torrey2009}, semi-supervised learning~\cite{ZhuRQK07}, and active learning~\cite{CastroKNQRZ08}. There is also analysis that connecting one-shot learning~\cite{Brenden2015} and Bayesian learning~\cite{tenenbaum2011grow} to human learning. It would be interesting to see how humans leverage their world knowledge when learning new problems or tasks.
\end{itemize}

The above open problems are still not discovered, and we regard them as equally important.
Moreover, as a new problem of machine learning, it would be interesting and important to have a general framework to set up machine learning algorithms when world knowledge is available.

\section*{Acknowledgments}

The authors wish to thank the anonymous reviewers of previous papers. We also thanks all the collaborators (Qiang Yang, Haixun Wang, Shusen Wang, Weizhu Chen, Zhongyuan Wang, Hongsong Li, Jiawei Han, Chenguang Wang, Ming Zhang, Shyam Upadhyay, Haoruo Peng, Stephen Mayhew, Xin Li) for the corresponding publications that have been referenced in the paper and new discussions and suggestions.

This work was supported by China 973 Fundamental R\&D Program (No.2014CB340304) and supported by DARPA under agreement numbers HR0011-15-2-0025. The U.S. Government is authorized to reproduce and distribute reprints for Governmental purposes notwithstanding any copyright notation thereon.  The views and conclusions contained herein are those of the authors and should not be interpreted as necessarily representing the official policies or endorsements, either expressed or implied, of any of the organizations that supported the work.

{
\bibliographystyle{IEEEtranS}
\bibliography{worldknowledgebib}

\begin{thebibliography}{100}
\providecommand{\url}[1]{#1}
\csname url@samestyle\endcsname
\providecommand{\newblock}{\relax}
\providecommand{\bibinfo}[2]{#2}
\providecommand{\BIBentrySTDinterwordspacing}{\spaceskip=0pt\relax}
\providecommand{\BIBentryALTinterwordstretchfactor}{4}
\providecommand{\BIBentryALTinterwordspacing}{\spaceskip=\fontdimen2\font plus
\BIBentryALTinterwordstretchfactor\fontdimen3\font minus
  \fontdimen4\font\relax}
\providecommand{\BIBforeignlanguage}[2]{{%
\expandafter\ifx\csname l@#1\endcsname\relax
\typeout{** WARNING: IEEEtranS.bst: No hyphenation pattern has been}%
\typeout{** loaded for the language `#1'. Using the pattern for}%
\typeout{** the default language instead.}%
\else
\language=\csname l@#1\endcsname
\fi
#2}}
\providecommand{\BIBdecl}{\relax}
\BIBdecl

\bibitem{AhnCPB16}
S.~Ahn, H.~Choi, T.~P{\"{a}}rnamaa, and Y.~Bengio, ``A neural knowledge
  language model,'' \emph{CoRR}, vol. abs/1608.00318, 2016.

\bibitem{AnastacioMC11}
I.~Anast{\'{a}}cio, B.~Martins, and P.~Calado, ``Supervised learning for
  linking named entities to knowledge base entries,'' in \emph{TAC}, 2011.

\bibitem{AngeliTWM14}
G.~Angeli, J.~Tibshirani, J.~Wu, and C.~D. Manning, ``Combining distant and
  partial supervision for relation extraction,'' in \emph{EMNLP}, 2014, pp.
  1556--1567.

\bibitem{ArtziFZ13}
Y.~Artzi, N.~FitzGerald, and L.~S. Zettlemoyer, ``Semantic parsing with
  combinatory categorial grammars,'' in \emph{ACL Tutorial Abstracts}, 2013,
  p.~2.

\bibitem{auer2007dbpedia}
S.~Auer, C.~Bizer, G.~Kobilarov, J.~Lehmann, R.~Cyganiak, and Z.~Ives,
  \emph{{DBpedia}: A nucleus for a web of open data}.\hskip 1em plus 0.5em
  minus 0.4em\relax Springer, 2007.

\bibitem{AytarZ11}
Y.~Aytar and A.~Zisserman, ``Tabula rasa: Model transfer for object category
  detection,'' in \emph{ICCV}, 2011, pp. 2252--2259.

\bibitem{BankoCSBE07}
M.~Banko, M.~J. Cafarella, S.~Soderland, M.~Broadhead, and O.~Etzioni, ``Open
  information extraction from the web,'' in \emph{IJCAI}, 2007, pp. 2670--2676.

\bibitem{basu2004probabilistic}
S.~Basu, M.~Bilenko, and R.~J. Mooney, ``A probabilistic framework for
  semi-supervised clustering,'' in \emph{KDD}, 2004, pp. 59--68.

\bibitem{Basu08}
S.~Basu, I.~Davidson, and K.~Wagstaff, \emph{Constrained Clustering: Advances
  in Algorithms, Theory, and Applications}.\hskip 1em plus 0.5em minus
  0.4em\relax Chapman \& Hall/CRC, 2008.

\bibitem{BelkinNS06}
M.~Belkin, P.~Niyogi, and V.~Sindhwani, ``Manifold regularization: {A}
  geometric framework for learning from labeled and unlabeled examples,''
  \emph{Journal of Machine Learning Research}, vol.~7, pp. 2399--2434, 2006.

\bibitem{Bengio09}
Y.~Bengio, ``Learning deep architectures for {AI},'' \emph{Foundations and
  Trends in Machine Learning}, vol.~2, no.~1, pp. 1--127, 2009.

\bibitem{BengioDVJ03}
Y.~Bengio, R.~Ducharme, P.~Vincent, and C.~Janvin, ``A neural probabilistic
  language model,'' \emph{Journal of Machine Learning Research}, vol.~3, pp.
  1137--1155, 2003.

\bibitem{BengioLPL06}
Y.~Bengio, P.~Lamblin, D.~Popovici, and H.~Larochelle, ``Greedy layer-wise
  training of deep networks,'' in \emph{NIPS}, 2006, pp. 153--160.

\bibitem{berant2013semantic}
J.~Berant, A.~Chou, R.~Frostig, and P.~Liang, ``Semantic parsing on freebase
  from question-answer pairs,'' in \emph{EMNLP}, 2013, pp. 1533--1544.

\bibitem{BerantL14}
J.~Berant and P.~Liang, ``Semantic parsing via paraphrasing,'' in \emph{ACL},
  2014, pp. 1415--1425.

\bibitem{BerantL15}
------, ``Imitation learning of agenda-based semantic parsers,'' \emph{{TACL}},
  vol.~3, pp. 545--558, 2015.

\bibitem{Blei03}
D.~M. Blei, A.~Y. Ng, and M.~I. Jordan, ``Latent {Dirichlet} allocation,''
  \emph{Journal of Machine Learning Research}, vol.~3, pp. 993--1022, 2003.

\bibitem{freebase}
K.~D. Bollacker, C.~Evans, P.~Paritosh, T.~Sturge, and J.~Taylor, ``Freebase: a
  collaboratively created graph database for structuring human knowledge,'' in
  \emph{SIGMOD}, 2008, pp. 1247--1250.

\bibitem{bragg2013crowdsourcing}
J.~Bragg, Mausam, and D.~S. Weld, ``Crowdsourcing multi-label classification
  for taxonomy creation,'' in \emph{{AAAI} Conference on Human Computation and
  Crowdsourcing ({HCOMP})}, 2013.

\bibitem{BrownPdLM92}
P.~F. Brown, V.~J.~D. Pietra, P.~V. de~Souza, J.~C. Lai, and R.~L. Mercer,
  ``Class-based n-gram models of natural language,'' \emph{Computational
  Linguistics}, vol.~18, no.~4, pp. 467--479, 1992.

\bibitem{BudanitskyH06}
A.~Budanitsky and G.~Hirst, ``Evaluating wordnet-based measures of lexical
  semantic relatedness,'' \emph{Computational Linguistics}, vol.~32, no.~1, pp.
  13--47, 2006.

\bibitem{BunescuM07}
R.~C. Bunescu and R.~J. Mooney, ``Learning to extract relations from the web
  using minimal supervision,'' in \emph{ACL}, 2007.

\bibitem{CaiY13}
Q.~Cai and A.~Yates, ``Large-scale semantic parsing via schema matching and
  lexicon extension,'' in \emph{ACL}, 2013, pp. 423--433.

\bibitem{CambriaSWH11}
E.~Cambria, Y.~Song, H.~Wang, and A.~Hussain, ``Isanette: {A} common and common
  sense knowledge base for opinion mining,'' in \emph{Data Mining Workshops
  (ICDMW)}, 2011, pp. 315--322.

\bibitem{NELL}
A.~Carlson, J.~Betteridge, B.~Kisiel, B.~Settles, E.~R.~H. Jr., and T.~M.
  Mitchell, ``Toward an architecture for never-ending language learning,'' in
  \emph{AAAI}, 2010, pp. 1306--1313.

\bibitem{CassidyCAJDRHRZ11}
T.~Cassidy, Z.~Chen, J.~Artiles, H.~Ji, H.~Deng, L.~Ratinov, J.~Han, D.~Roth,
  and J.~Zheng, ``{CUNY-UIUC-SRI} {TAC-KBP2011} entity linking system
  description,'' in \emph{TAC}, 2011.

\bibitem{CassidyJDZH12}
T.~Cassidy, H.~Ji, H.~Deng, J.~Zheng, and J.~Han, ``Analysis and refinement of
  cross-lingual entity linking,'' in \emph{Information Access Evaluation.
  Multilinguality, Multimodality, and Visual Analytics - Third International
  Conference of the {CLEF} Initiative}, 2012, pp. 1--12.

\bibitem{CassidyJRZH12}
T.~Cassidy, H.~Ji, L.~Ratinov, A.~Zubiaga, and H.~Huang, ``Analysis and
  enhancement of wikification for microblogs with context expansion,'' in
  \emph{COLING}, 2012, pp. 441--456.

\bibitem{CastroKNQRZ08}
R.~M. Castro, C.~Kalish, R.~D. Nowak, R.~Qian, T.~T. Rogers, and X.~Zhu,
  ``Human active learning,'' in \emph{NIPS}, 2008, pp. 241--248.

\bibitem{CeccarelliLOPT13}
D.~Ceccarelli, C.~Lucchese, S.~Orlando, R.~Perego, and S.~Trani, ``Learning
  relatedness measures for entity linking,'' in \emph{CIKM}, 2013, pp.
  139--148.

\bibitem{ChangSR13}
K.~Chang, R.~Samdani, and D.~Roth, ``A constrained latent variable model for
  coreference resolution,'' in \emph{EMNLP}, 2013, pp. 601--612.

\bibitem{ChangLRS2008}
M.-W. Chang, L.~Ratinov, D.~Roth, and V.~Srikumar, ``Importance of semantic
  representation: Dataless classification,'' in \emph{AAAI}, 2008, pp.
  830--835.

\bibitem{ChangRR12}
M.-W. Chang, L.-A. Ratinov, and D.~Roth, ``Structured learning with constrained
  conditional models,'' \emph{Machine Learning}, vol.~88, no.~3, pp. 399--431,
  2012.

\bibitem{ChangSGR10}
M.~Chang, V.~Srikumar, D.~Goldwasser, and D.~Roth, ``Structured output learning
  with indirect supervision,'' in \emph{ICML}, 2010, pp. 199--206.

\bibitem{ChaSchZie06}
O.~Chapelle, B.~Sch{\"o}lkopf, and A.~Zien, Eds., \emph{Semi-Supervised
  Learning}.\hskip 1em plus 0.5em minus 0.4em\relax MIT Press, 2006.

\bibitem{0001LZC16}
J.~Chen, K.~Li, J.~Zhu, and W.~Chen, ``Warplda: a cache efficient {O(1)}
  algorithm for latent dirichlet allocation,'' \emph{{PVLDB}}, vol.~9, no.~10,
  pp. 744--755, 2016.

\bibitem{ChenJ11}
Z.~Chen and H.~Ji, ``Collaborative ranking: {A} case study on entity linking,''
  in \emph{EMNLP}, 2011, pp. 771--781.

\bibitem{ChenTLLLSAPJ10}
Z.~Chen, S.~Tamang, A.~Lee, X.~Li, W.~Lin, M.~G. Snover, J.~Artiles,
  M.~Passantino, and H.~Ji, ``{CUNY-BLENDER} {TAC-KBP2010} entity linking and
  slot filling system description,'' in \emph{TAC}, 2010.

\bibitem{ChengR13}
X.~Cheng and D.~Roth, ``Relational inference for wikification,'' in
  \emph{EMNLP}, 2013, pp. 1787--1796.

\bibitem{chilton2013cascade}
L.~B. Chilton, G.~Little, D.~Edge, D.~S. Weld, and J.~A. Landay, ``Cascade:
  Crowdsourcing taxonomy creation,'' in \emph{SIGCHI Conference on Human
  Factors in Computing Systems}.\hskip 1em plus 0.5em minus 0.4em\relax ACM,
  2013, pp. 1999--2008.

\bibitem{cimiano2003}
P.~Cimiano, S.~Staab, and J.~Tane, ``Automatic acquisition of taxonomies from
  text: {FCA} meets {NLP},'' in \emph{ECML/PKDD Workshop on Adaptive Text
  Extraction and Mining, Cavtat-Dubrovnik, Croatia}, 2003, pp. 10--17.

\bibitem{Cissetel01142046}
\BIBentryALTinterwordspacing
M.~M. Cisse, ``{Efficient extreme classification},'' Theses, {Universit{\'e}
  Pierre et Marie Curie - Paris VI}, Jul. 2014. [Online]. Available:
  \url{https://tel.archives-ouvertes.fr/tel-01142046}
\BIBentrySTDinterwordspacing

\bibitem{ClarkeGCR10}
J.~Clarke, D.~Goldwasser, M.~Chang, and D.~Roth, ``Driving semantic parsing
  from the world's response,'' in \emph{CoNLL}, 2010, pp. 18--27.

\bibitem{Collobert11}
R.~Collobert, J.~Weston, L.~Bottou, M.~Karlen, K.~Kavukcuoglu, and P.~P. Kuksa,
  ``Natural language processing (almost) from scratch,'' \emph{Journal of
  Machine Learning Research}, vol.~12, pp. 2493--2537, 2011.

\bibitem{Cunningham2015}
J.~P. Cunningham and Z.~Ghahramani, ``Linear dimensionality reduction: survey,
  insights, and generalizations,'' \emph{Journal of Machine Learning Research},
  vol.~16, pp. 3619--3622, 2015.

\bibitem{DaumeLM09}
H.~{Daum{\'{e}} III}, J.~Langford, and D.~Marcu, ``Search-based structured
  prediction,'' \emph{Machine Learning}, vol.~75, no.~3, pp. 297--325, 2009.

\bibitem{Davis2015}
E.~Davis and G.~Marcus, ``Commonsense reasoning and commonsense knowledge in
  artificial intelligence,'' \emph{Commun. ACM}, vol.~58, no.~9, pp. 92--103,
  Aug. 2015.

\bibitem{deerwester1990indexing}
S.~Deerwester, S.~Dumais, G.~Furnas, T.~Landauer, and R.~Harshman, ``Indexing
  by latent semantic analysis,'' \emph{Journal of the American society for
  information science}, vol.~41, no.~6, pp. 391--407, 1990.

\bibitem{dong2014knowledge}
X.~Dong, E.~Gabrilovich, G.~Heitz, W.~Horn, N.~Lao, K.~Murphy, T.~Strohmann,
  S.~Sun, and W.~Zhang, ``Knowledge vault: A web-scale approach to
  probabilistic knowledge fusion,'' in \emph{KDD}, 2014, pp. 601--610.

\bibitem{DredzeMRGF10}
M.~Dredze, P.~McNamee, D.~Rao, A.~Gerber, and T.~Finin, ``Entity disambiguation
  for knowledge base population,'' in \emph{COLING}, 2010, pp. 277--285.

\bibitem{Egozi2011}
O.~Egozi, S.~Markovitch, and E.~Gabrilovich, ``Concept-based information
  retrieval using explicit semantic analysis,'' \emph{ACM Transactions on
  Information Systems (TOIS)}, vol.~29, no.~2, pp. 8:1--8:34, 2011.

\bibitem{Elhoseiny13}
M.~Elhoseiny, B.~Saleh, and A.~M. Elgammal, ``Write a classifier: Zero shot
  learning using purely textual descriptions,'' in \emph{ICCV}, 2013, pp.
  2584--2591.

\bibitem{knowitall}
O.~Etzioni, M.~Cafarella, and D.~Downey, ``Webscale information extraction in
  knowitall (preliminary results),'' in \emph{WWW}, 2004, pp. 100--110.

\bibitem{FaderZE13}
A.~Fader, L.~S. Zettlemoyer, and O.~Etzioni, ``Paraphrase-driven learning for
  open question answering,'' in \emph{ACLs}, 2013, pp. 1608--1618.

\bibitem{FanZZLZC14}
M.~Fan, D.~Zhao, Q.~Zhou, Z.~Liu, T.~F. Zheng, and E.~Y. Chang, ``Distant
  supervision for relation extraction with matrix completion,'' in \emph{ACL},
  2014, pp. 839--849.

\bibitem{FangZWCL16}
W.~Fang, J.~Zhang, D.~Wang, Z.~Chen, and M.~Li, ``Entity disambiguation by
  knowledge and text jointly embedding,'' in \emph{CoNLL}, 2016, pp. 260--269.

\bibitem{FangLZWCL16}
Y.~Fang, W.~Lin, V.~W. Zheng, M.~Wu, K.~C. Chang, and X.~Li, ``Semantic
  proximity search on graphs with metagraph-based learning,'' in \emph{ICDE},
  2016, pp. 277--288.

\bibitem{Faure98}
D.~Faure and C.~N\'{e}dellec, ``A corpus-based conceptual clustering method for
  verb frames and ontology acquisition,'' in \emph{In {LREC} workshop on
  adapting lexical and corpus resources to sublanguages and applications},
  1998, pp. 5--12.

\bibitem{FaureN99}
------, ``Knowledge acquisition of predicate argument structures from technical
  texts using machine learning: The system {ASIUM},'' in \emph{EKAW}, 1999, pp.
  329--334.

\bibitem{Zhiye2015}
Z.~Fei, D.~Khashabi, H.~Peng, H.~Wu, and D.~Roth, ``Coreference resolution with
  world knowledge,'' in \emph{IJCAI Workshop on Cognitive Knowledge Acquisition
  and Applications}, 2015, pp. 814--824.

\bibitem{wordnet}
C.~Fellbaum, Ed., \emph{WordNet: an electronic lexical database}.\hskip 1em
  plus 0.5em minus 0.4em\relax MIT Press, 1998.

\bibitem{FerraginaS10}
P.~Ferragina and U.~Scaiella, ``{TAGME:} on-the-fly annotation of short text
  fragments (by wikipedia entities),'' in \emph{CIKM}, 2010, pp. 1625--1628.

\bibitem{FlorianJKZ06}
R.~Florian, H.~Jing, N.~Kambhatla, and I.~Zitouni, ``Factorizing complex
  models: {A} case study in mention detection,'' in \emph{ACL}, 2006.

\bibitem{Fodeh2011}
S.~Fodeh, B.~Punch, and P.-N. Tan, ``On ontology-driven document clustering
  using core semantic features,'' \emph{Knowl. Inf. Syst.}, vol.~28, no.~2, pp.
  395--421, 2011.

\bibitem{GabrilovichM05}
E.~Gabrilovich and S.~Markovitch, ``Feature generation for text categorization
  using world knowledge,'' in \emph{IJCAI}, 2005, pp. 1048--1053.

\bibitem{GabrilovichM06}
------, ``Overcoming the brittleness bottleneck using {Wikipedia}: Enhancing
  text categorization with encyclopedic knowledge,'' in \emph{AAAI}, 2006, pp.
  1301--1306.

\bibitem{GabrilovichM07}
------, ``Computing semantic relatedness using {Wikipedia-based} explicit
  semantic analysis,'' in \emph{IJCAI}, 2007, pp. 1606--1611.

\bibitem{GabrilovichM09}
------, ``Wikipedia-based semantic interpretation for natural language
  processing,'' \emph{Journal of Artificial Intelligence Research {(JAIR)}},
  vol.~34, no.~1, pp. 443--498, 2009.

\bibitem{Gabrilovich13FreebaseClueWeb}
E.~Gabrilovich, M.~Ringgaard, and A.~Subramanya, ``Facc1: Freebase annotation
  of clueweb corpora, version 1,'' in \emph{Release date 2013-06-26, Format
  version 1, Correction level 0}, 2013.

\bibitem{ganchev2010posterior}
K.~Ganchev, J.~Gra{\c{c}}a, J.~Gillenwater, and B.~Taskar, ``Posterior
  regularization for structured latent variable models,'' \emph{JMLR}, vol.~11,
  pp. 2001--2049, 2010.

\bibitem{GoldwasserRCR11}
D.~Goldwasser, R.~Reichart, J.~Clarke, and D.~Roth, ``Confidence driven
  unsupervised semantic parsing,'' in \emph{ACL}, 2011, pp. 1486--1495.

\bibitem{GopalY13}
S.~Gopal and Y.~Yang, ``Recursive regularization for large-scale classification
  with hierarchical and graphical dependencies,'' in \emph{The 19th {ACM}
  {SIGKDD} International Conference on Knowledge Discovery and Data Mining,
  {KDD} 2013, Chicago, IL, USA, August 11-14, 2013}, 2013, pp. 257--265.

\bibitem{Grefenstette92}
G.~Grefenstette, ``Sextant: Exploring unexplored contexts for semantic
  extraction from syntactic analysis,'' in \emph{ACL}, 1992, pp. 324--326.

\bibitem{Griffiths04}
T.~L. Griffiths and M.~Steyvers, ``Finding scientific topics,''
  \emph{Proceedings of the National Academy of Sciences}, vol. 101, pp.
  5228--5235, 2004.

\bibitem{Grishman1994}
R.~Grishman and J.~Sterling, ``Generalizing automatically generated selectional
  patterns,'' in \emph{COLING}, 1994, pp. 742--747.

\bibitem{GuoWSH14}
X.~Guo, H.~Wang, Y.~Song, and G.~Hong, ``Brief survey of crowdsourcing for data
  mining,'' \emph{Expert Syst. Appl.}, vol.~41, no.~17, pp. 7987--7994, 2014.

\bibitem{Ha-Thuc11}
V.~Ha-Thuc and J.-M. Renders, ``Large-scale hierarchical text classification
  without labelled data,'' in \emph{WSDM}, 2011, pp. 685--694.

\bibitem{HacheyRC11}
B.~Hachey, W.~Radford, and J.~R. Curran, ``Graph-based named entity linking
  with wikipedia,'' in \emph{WISE}, 2011, pp. 213--226.

\bibitem{HakimovOD12}
S.~Hakimov, S.~A. Oto, and E.~Dogdu, ``Named entity recognition and
  disambiguation using linked data and graph-based centrality scoring,'' in
  \emph{SWIM}, 2012, p.~4.

\bibitem{Hindle1990}
D.~Hindle, ``Noun classification from predicate-argument structures,'' in
  \emph{ACL}, 1990, pp. 268--275.

\bibitem{HinSal06}
G.~Hinton and R.~Salakhutdinov, ``Reducing the dimensionality of data with
  neural networks,'' \emph{Science}, vol. 313, no. 5786, pp. 504--507, 2006.

\bibitem{HintonOT06}
G.~E. Hinton, S.~Osindero, and Y.~W. Teh, ``A fast learning algorithm for deep
  belief nets,'' \emph{Neural Computation}, vol.~18, no.~7, pp. 1527--1554,
  2006.

\bibitem{HochreiterS97}
S.~Hochreiter and J.~Schmidhuber, ``Long short-term memory,'' \emph{Neural
  Computation}, vol.~9, no.~8, pp. 1735--1780, 1997.

\bibitem{HoffartYBFPSTTW11}
J.~Hoffart, M.~A. Yosef, I.~Bordino, H.~F{\"{u}}rstenau, M.~Pinkal, M.~Spaniol,
  B.~Taneva, S.~Thater, and G.~Weikum, ``Robust disambiguation of named
  entities in text,'' in \emph{EMNLP}, 2011, pp. 782--792.

\bibitem{HoffmannZLZW11}
R.~Hoffmann, C.~Zhang, X.~Ling, L.~S. Zettlemoyer, and D.~S. Weld,
  ``Knowledge-based weak supervision for information extraction of overlapping
  relations,'' in \emph{ACL}, 2011, pp. 541--550.

\bibitem{Hofmann99}
T.~Hofmann, ``Probabilistic latent semantic analysis,'' in \emph{UAI}, 1999,
  pp. 289--296.

\bibitem{HuJian2008}
J.~Hu, L.~Fang, Y.~Cao, H.-J. Zeng, H.~Li, Q.~Yang, and Z.~Chen, ``Enhancing
  text clustering by leveraging {Wikipedia} semantics,'' in \emph{SIGIR}, 2008,
  pp. 179--186.

\bibitem{Hu2009EIE}
X.~Hu, N.~Sun, C.~Zhang, and T.-S. Chua, ``Exploiting internal and external
  semantics for the clustering of short texts using world knowledge,'' in
  \emph{CIKM}, 2009, pp. 919--928.

\bibitem{HuXiaohua2009}
X.~Hu, X.~Zhang, C.~Lu, E.~K. Park, and X.~Zhou, ``Exploiting wikipedia as
  external knowledge for document clustering,'' in \emph{KDD}, 2009, pp.
  389--396.

\bibitem{HuMLHX16}
Z.~Hu, X.~Ma, Z.~Liu, E.~H. Hovy, and E.~P. Xing, ``Harnessing deep neural
  networks with logic rules,'' in \emph{ACL}, 2016.

\bibitem{Hua2013}
W.~Hua, Y.~Song, H.~Wang, and X.~Zhou, ``Identifying users' topical tasks in
  {W}eb search,'' in \emph{WSDM}, 2013, pp. 93--102.

\bibitem{HuangCHJL14}
H.~Huang, Y.~Cao, X.~Huang, H.~Ji, and C.~Lin, ``Collective tweet wikification
  based on semi-supervised graph regularization,'' in \emph{ACL}, 2014, pp.
  380--390.

\bibitem{HuangZCSML16}
Z.~Huang, Y.~Zheng, R.~Cheng, Y.~Sun, N.~Mamoulis, and X.~Li, ``Meta structure:
  Computing relevance in large heterogeneous information networks,'' in
  \emph{SIGKDD}, 2016, pp. 1595--1604.

\bibitem{JiG11}
H.~Ji and R.~Grishman, ``Knowledge base population: Successful approaches and
  challenges,'' in \emph{ACL-HLT}, 2011, pp. 1148--1158.

\bibitem{HeJiang2017}
H.~Jiang, Y.~Song, C.~Wang, M.~Zhang, and Y.~Sun, ``Semi-supervised learning
  over heterogeneous information networks by ensemble of meta-graph guided
  random walks,'' in \emph{IJCAI}, 2017.

\bibitem{JohnsonKSLSBL15}
J.~Johnson, R.~Krishna, M.~Stark, L.~Li, D.~A. Shamma, M.~S. Bernstein, and
  F.~Li, ``Image retrieval using scene graphs,'' in \emph{{CVPR}}, 2015, pp.
  3668--3678.

\bibitem{JohnsonSLKWCTVW16}
M.~Johnson, M.~Schuster, Q.~V. Le, M.~Krikun, Y.~Wu, Z.~Chen, N.~Thorat, F.~B.
  Vi{\'{e}}gas, M.~Wattenberg, G.~Corrado, M.~Hughes, and J.~Dean, ``Google's
  multilingual neural machine translation system: Enabling zero-shot
  translation,'' \emph{CoRR}, vol. abs/1611.04558, 2016.

\bibitem{jurgens-pilehvar:2015:EMNLP-Tutorials}
D.~Jurgens and M.~T. Pilehvar, ``Semantic similarity frontiers: From concepts
  to documents,'' in \emph{EMNLP Tutorials}, 2015, pp. 1--2.

\bibitem{Kim2013}
D.~Kim, H.~Wang, and A.~Oh, ``Context-dependent conceptualization,'' in
  \emph{IJCAI}, 2013, pp. 2654--2661.

\bibitem{KollerFriedman09}
D.~Koller and N.~Friedman, \emph{Probabilistic Graphical Models: Principles and
  Techniques}.\hskip 1em plus 0.5em minus 0.4em\relax MIT Press, 2009.

\bibitem{KotlermanDSZ10}
L.~Kotlerman, I.~Dagan, I.~Szpektor, and M.~Zhitomirsky{-}Geffet, ``Directional
  distributional similarity for lexical inference,'' \emph{Natural Language
  Engineering}, vol.~16, no.~4, pp. 359--389, 2010.

\bibitem{Kozareva2011}
Z.~Kozareva and S.~Ravi, ``Unsupervised name ambiguity resolution using a
  generative model,'' in \emph{EMNLP Workshop on Unsupervised Learning in NLP},
  2011.

\bibitem{KrishnaZGJHKCKL16}
\BIBentryALTinterwordspacing
R.~Krishna, Y.~Zhu, O.~Groth, J.~Johnson, K.~Hata, J.~Kravitz, S.~Chen,
  Y.~Kalantidis, L.~Li, D.~A. Shamma, M.~S. Bernstein, and F.~Li, ``Visual
  genome: Connecting language and vision using crowdsourced dense image
  annotations,'' \emph{CoRR}, vol. abs/1602.07332, 2016. [Online]. Available:
  \url{http://arxiv.org/abs/1602.07332}
\BIBentrySTDinterwordspacing

\bibitem{Krishnamurthy2012}
J.~Krishnamurthy and T.~M. Mitchell, ``Weakly supervised training of semantic
  parsers,'' in \emph{EMNLP-CoNLL}, 2012, pp. 754--765.

\bibitem{KulkarniSRC09}
S.~Kulkarni, A.~Singh, G.~Ramakrishnan, and S.~Chakrabarti, ``Collective
  annotation of wikipedia entities in web text,'' in \emph{SIGKDD}, 2009, pp.
  457--466.

\bibitem{KwiatkowskiCAZ13}
T.~Kwiatkowski, E.~Choi, Y.~Artzi, and L.~S. Zettlemoyer, ``Scaling semantic
  parsers with on-the-fly ontology matching,'' in \emph{EMNLP}, 2013, pp.
  1545--1556.

\bibitem{KwiatkowksiZGS10}
T.~Kwiatkowski, L.~S. Zettlemoyer, S.~Goldwater, and M.~Steedman, ``Inducing
  probabilistic {CCG} grammars from logical form with higher-order
  unification,'' in \emph{EMNLP}, 2010, pp. 1223--1233.

\bibitem{KwiatkowskiZGS11}
------, ``Lexical generalization in {CCG} grammar induction for semantic
  parsing,'' in \emph{EMNLP}, 2011, pp. 1512--1523.

\bibitem{Brenden2015}
B.~M. Lake, R.~Salakhutdinov, and J.~B. Tenenbaum, ``Human-level concept
  learning through probabilistic program induction,'' \emph{Science}, vol. 350,
  no. 6266, pp. 1332--1338, Dec. 2015.

\bibitem{LaoC10}
N.~Lao and W.~W. Cohen, ``Relational retrieval using a combination of
  path-constrained random walks,'' \emph{Machine Learning}, vol.~81, no.~1, pp.
  53--67, 2010.

\bibitem{LarochelleECBB07}
H.~Larochelle, D.~Erhan, A.~C. Courville, J.~Bergstra, and Y.~Bengio, ``An
  empirical evaluation of deep architectures on problems with many factors of
  variation,'' in \emph{ICML}, 2007, pp. 473--480.

\bibitem{LazaridouDB15}
A.~Lazaridou, G.~Dinu, and M.~Baroni, ``Hubness and pollution: Delving into
  cross-space mapping for zero-shot learning,'' in \emph{ACL}, 2015, pp.
  270--280.

\bibitem{Lease11}
M.~Lease, ``On quality control and machine learning in crowdsourcing,'' in
  \emph{Human Computation}, 2011.

\bibitem{YannLeCun15}
Y.~LeCun, Y.~Bengio, and G.~Hinton, ``Deep learning,'' \emph{Nature}, vol. 521,
  pp. 436--444, 2015.

\bibitem{Lee2013DCR}
H.~Lee, A.~Chang, Y.~Peirsman, N.~Chambers, M.~Surdeanu, and D.~Jurafsky,
  ``Deterministic coreference resolution based on entity-centric,
  precision-ranked rules,'' \emph{Comput. Linguist.}, vol.~39, no.~4, pp.
  885--916, 2013.

\bibitem{LeeRTN09}
H.~Lee, R.~Raina, A.~Teichman, and A.~Y. Ng, ``Exponential family sparse coding
  with application to self-taught learning,'' in \emph{IJCAI}, 2009, pp.
  1113--1119.

\bibitem{researchCyc}
D.~B. Lenat and R.~V. Guha, \emph{Building Large Knowledge-Based Systems:
  Representation and Inference in the Cyc Project}.\hskip 1em plus 0.5em minus
  0.4em\relax Addison-Wesley, 1989.

\bibitem{Levesque11}
H.~J. Levesque, ``The winograd schema challenge,'' in \emph{Logical
  Formalizations of Commonsense Reasoning, {AAAI} Spring Symposium}, 2011.

\bibitem{LevyG14}
O.~Levy and Y.~Goldberg, ``Dependency-based word embeddings,'' in \emph{ACL},
  2014, pp. 302--308.

\bibitem{LevyGD15}
O.~Levy, Y.~Goldberg, and I.~Dagan, ``Improving distributional similarity with
  lessons learned from word embeddings,'' \emph{{TACL}}, vol.~3, pp. 211--225,
  2015.

\bibitem{LiARS14}
A.~Q. Li, A.~Ahmed, S.~Ravi, and A.~J. Smola, ``Reducing the sampling
  complexity of topic models,'' in \emph{KDD}, 2014, pp. 891--900.

\bibitem{FeiFeiFP06}
F.~Li, R.~Fergus, and P.~Perona, ``One-shot learning of object categories,''
  \emph{{IEEE} Trans. Pattern Anal. Mach. Intell.}, vol.~28, no.~4, pp.
  594--611, 2006.

\bibitem{LiJ14}
Q.~Li and H.~Ji, ``Incremental joint extraction of entity mentions and
  relations,'' in \emph{ACL}, 2014, pp. 402--412.

\bibitem{LiZTHIS16}
Y.~Li, R.~Zheng, T.~Tian, Z.~Hu, R.~Iyer, and K.~P. Sycara, ``Joint embedding
  of hierarchical categories and entities for concept categorization and
  dataless classification,'' in \emph{COLING}, 2016, pp. 2678--2688.

\bibitem{PLiang05}
P.~Liang, ``Semi-supervised learning for natural language,'' Master's thesis,
  Massachusetts Institute of Technology, 2005.

\bibitem{Liang13}
------, ``Lambda dependency-based compositional semantics,'' \emph{arXiv},
  2013.

\bibitem{LiangJK09}
P.~Liang, M.~I. Jordan, and D.~Klein, ``Learning from measurements in
  exponential families,'' in \emph{ICML}, 2009, pp. 641--648.

\bibitem{LiangJK11}
------, ``Learning dependency-based compositional semantics,'' in \emph{ACL},
  2011, pp. 590--599.

\bibitem{FangzhenLin20103}
F.~Lin, ``From satisfiability to linear algebra,'' in \emph{Technical
  Report}.\hskip 1em plus 0.5em minus 0.4em\relax HKUST, 2013.

\bibitem{LinSLLS16}
Y.~Lin, S.~Shen, Z.~Liu, H.~Luan, and M.~Sun, ``Neural relation extraction with
  selective attention over instances,'' in \emph{ACL}, 2016.

\bibitem{LiuFSK1998}
H.~Liu and H.~Motoda, \emph{Feature Selection for Knowledge Discovery and Data
  Mining}.\hskip 1em plus 0.5em minus 0.4em\relax Kluwer Academic Publishers,
  1998.

\bibitem{LiuYWZCM05}
T.~Liu, Y.~Yang, H.~Wan, H.~Zeng, Z.~Chen, and W.~Ma, ``Support vector machines
  classification with a very large-scale taxonomy,'' \emph{{SIGKDD}
  Explorations}, vol.~7, no.~1, pp. 36--43, 2005.

\bibitem{Liu2012}
X.~Liu, Y.~Song, S.~Liu, and H.~Wang, ``Automatic taxonomy construction from
  keywords,'' in \emph{KDD}, 2012, pp. 1433--1441.

\bibitem{LiuLXZ14}
Y.~Liu, K.~Liu, L.~Xu, and J.~Zhao, ``Exploring fine-grained entity type
  constraints for distantly supervised relation extraction,'' in
  \emph{{COLING}}, 2014, pp. 2107--2116.

\bibitem{LuL04}
Z.~Lu and T.~K. Leen, ``Semi-supervised learning with penalized probabilistic
  clustering,'' in \emph{NIPS}, 2004, pp. 849--856.

\bibitem{LuZPXW014}
Z.~Lu, Y.~Zhu, S.~J. Pan, E.~W. Xiang, Y.~Wang, and Q.~Yang, ``Source free
  transfer learning for text classification,'' in \emph{AAAI}, 2014, pp.
  122--128.

\bibitem{MannM10}
G.~S. Mann and A.~McCallum, ``Generalized expectation criteria for
  semi-supervised learning with weakly labeled data,'' \emph{Journal of Machine
  Learning Research}, vol.~11, pp. 955--984, 2010.

\bibitem{Manning2008IIR}
C.~D. Manning, P.~Raghavan, and H.~Sch\"{u}tze, \emph{Introduction to
  Information Retrieval}.\hskip 1em plus 0.5em minus 0.4em\relax New York, NY,
  USA: Cambridge University Press, 2008.

\bibitem{MedelyanMLW09}
O.~Medelyan, D.~N. Milne, C.~Legg, and I.~H. Witten, ``Mining meaning from
  wikipedia,'' \emph{Int. J. Hum.-Comput. Stud.}, vol.~67, no.~9, pp. 716--754,
  2009.

\bibitem{MeijWR12}
E.~Meij, W.~Weerkamp, and M.~de~Rijke, ``Adding semantics to microblog posts,''
  in \emph{WSDM}, 2012, pp. 563--572.

\bibitem{MendesDRSB12}
P.~N. Mendes, J.~Daiber, R.~K. Rajapakse, F.~Sasaki, and C.~Bizer, ``Evaluating
  the impact of phrase recognition on concept tagging,'' in \emph{LREC}, 2012,
  pp. 1277--1280.

\bibitem{MendesJGB11}
P.~N. Mendes, M.~Jakob, A.~Garc{\'{\i}}a{-}Silva, and C.~Bizer, ``Dbpedia
  spotlight: shedding light on the web of documents,'' in \emph{I-SEMANTICS},
  2011, pp. 1--8.

\bibitem{MengT0C15}
R.~Meng, Y.~Tong, L.~Chen, and C.~C. Cao, ``Crowdtc: Crowdsourced taxonomy
  construction,'' in \emph{ICDM}, 2015, pp. 913--918.

\bibitem{Mikolov133}
\BIBentryALTinterwordspacing
T.~Mikolov, K.~Chen, G.~Corrado, and J.~Dean, ``Efficient estimation of word
  representations in vector space,'' \emph{ICLR}, 2013. [Online]. Available:
  \url{http://arxiv.org/abs/1301.3781}
\BIBentrySTDinterwordspacing

\bibitem{Mikolov132}
T.~Mikolov, I.~Sutskever, K.~Chen, G.~S. Corrado, and J.~Dean, ``Distributed
  representations of words and phrases and their compositionality,'' in
  \emph{NIPS}, 2013, pp. 3111--3119.

\bibitem{MikolovYZ13}
T.~Mikolov, W.-t. Yih, and G.~Zweig, ``Linguistic regularities in continuous
  space word representations,'' in \emph{NAACL-HLT}, 2013, pp. 746--751.

\bibitem{MilneW08}
D.~N. Milne and I.~H. Witten, ``Learning to link with wikipedia,'' in
  \emph{CIKM}, 2008, pp. 509--518.

\bibitem{MinGW0G13}
B.~Min, R.~Grishman, L.~Wan, C.~Wang, and D.~Gondek, ``Distant supervision for
  relation extraction with an incomplete knowledge base,'' in \emph{HLT-NAACL},
  2013, pp. 777--782.

\bibitem{mintz2009distant}
M.~Mintz, S.~Bills, R.~Snow, and D.~Jurafsky, ``Distant supervision for
  relation extraction without labeled data,'' in \emph{ACL/AFNLP}, 2009, pp.
  1003--1011.

\bibitem{mitamura2015event}
T.~Mitamura, Y.~Yamakawa, S.~Holm, Z.~Song, A.~Bies, S.~Kulick, and
  S.~Strassel, ``Event nugget annotation: Processes and issues,'' in
  \emph{NAACL-HLT 3rd Workshop on EVENTS}, 2015, pp. 66--76.

\bibitem{MnihH08}
A.~Mnih and G.~E. Hinton, ``A scalable hierarchical distributed language
  model,'' in \emph{NIPS}, 2008, pp. 1081--1088.

\bibitem{Mooney07}
R.~J. Mooney, ``Learning for semantic parsing,'' in \emph{CICLing}, 2007, pp.
  311--324.

\bibitem{MorinB05}
F.~Morin and Y.~Bengio, ``Hierarchical probabilistic neural network language
  model,'' in \emph{AISTATS}, 2005.

\bibitem{Movshovitz-Attias15}
D.~Movshovitz{-}Attias and W.~W. Cohen, ``{KB-LDA:} jointly learning a
  knowledge base of hierarchy, relations, and facts,'' in \emph{ACL}, 2015, pp.
  1449--1459.

\bibitem{NageshHR14}
A.~Nagesh, G.~Haffari, and G.~Ramakrishnan, ``Noisy or-based model for relation
  extraction using distant supervision,'' in \emph{EMNLP}, 2014, pp.
  1937--1941.

\bibitem{NewmanASW09}
D.~Newman, A.~U. Asuncion, P.~Smyth, and M.~Welling, ``Distributed algorithms
  for topic models,'' \emph{Journal of Machine Learning Research}, vol.~10, pp.
  1801--1828, 2009.

\bibitem{Nickel0TG16}
M.~Nickel, K.~Murphy, V.~Tresp, and E.~Gabrilovich, ``A review of relational
  machine learning for knowledge graphs,'' \emph{Proceedings of the {IEEE}},
  vol. 104, no.~1, pp. 11--33, 2016.

\bibitem{NiepertD15}
M.~Niepert and P.~M. Domingos, ``Learning and inference in tractable
  probabilistic knowledge bases,'' in \emph{UAI}, 2015, pp. 632--641.

\bibitem{NigamMTM00}
K.~Nigam, A.~McCallum, S.~Thrun, and T.~M. Mitchell, ``Text classification from
  labeled and unlabeled documents using {EM},'' \emph{Machine Learning},
  vol.~39, no. 2/3, pp. 103--134, 2000.

\bibitem{NIST05}
NIST, ``The {ACE} evaluation plan.'' 2005.

\bibitem{PalatucciPHM09}
M.~Palatucci, D.~Pomerleau, G.~E. Hinton, and T.~M. Mitchell, ``Zero-shot
  learning with semantic output codes,'' in \emph{NIPS}, 2009, pp. 1410--1418.

\bibitem{2010Palmer}
M.~Palmer, D.~Gildea, and N.~Xue, \emph{Semantic Role Labeling}, ser. Synthesis
  Lectures on Human Language Technologies.\hskip 1em plus 0.5em minus
  0.4em\relax Morgan {\&} Claypool Publishers, 2010.

\bibitem{Pan2010}
S.~J. Pan and Q.~Yang, ``A survey on transfer learning,'' \emph{IEEE TKDE},
  vol.~22, no.~10, pp. 1345--1359, 2010.

\bibitem{ParkHW16}
J.~Park, S.~Hwang, and H.~Wang, ``Fine-grained semantic conceptualization of
  framenet,'' in \emph{AAAI}, 2016, pp. 2638--2644.

\bibitem{PengKhRo15}
H.~Peng, D.~Khashabi, and D.~Roth, ``Solving hard coreference problems,'' in
  \emph{NAACL-HLT}, 2015, pp. 809--819.

\bibitem{PengSR16}
H.~Peng, Y.~Song, and D.~Roth, ``Event detection and co-reference with minimal
  supervision,'' in \emph{EMNLP}, 2016, pp. 392--402.

\bibitem{PershinaMXG14}
M.~Pershina, B.~Min, W.~Xu, and R.~Grishman, ``Infusion of labeled data into
  distant supervision for relation extraction,'' in \emph{ACL}, 2014, pp.
  732--738.

\bibitem{PinkNRCNTC13}
G.~Pink, A.~Naoum, W.~Radford, W.~Cannings, J.~Nothman, D.~Tse, and J.~R.
  Curran, ``{SYDNEY} {CMCRC} at {TAC} 2013,'' in \emph{TAC}, 2013.

\bibitem{PonzettoS07}
S.~P. Ponzetto and M.~Strube, ``Deriving a large-scale taxonomy from
  wikipedia,'' in \emph{AAAI}, 2007, pp. 1440--1445.

\bibitem{Poon13}
H.~Poon, ``Grounded unsupervised semantic parsing,'' in \emph{ACL}, 2013, pp.
  933--943.

\bibitem{Porteous08}
I.~Porteous, D.~Newman, A.~T. Ihler, A.~Asuncion, P.~Smyth, and M.~Welling,
  ``Fast collapsed gibbs sampling for latent dirichlet allocation,'' in
  \emph{KDD}, 2008, pp. 569--577.

\bibitem{Potthast2008}
M.~Potthast, B.~Stein, and M.~Anderka, ``A wikipedia-based multilingual
  retrieval model,'' in \emph{ECIR}, 2008, pp. 522--530.

\bibitem{PunyakanokRY08}
V.~Punyakanok, D.~Roth, and W.~Yih, ``The importance of syntactic parsing and
  inference in semantic role labeling,'' \emph{Computational Linguistics},
  vol.~34, no.~2, pp. 257--287, 2008.

\bibitem{QiWuVTT2015}
H.~Qi, T.~Wu, M.~W. Lee, and S.-C. Zhu, ``A restricted visual turing test for
  deep scene and event understanding,'' 2015.

\bibitem{Rabiner1989}
L.~R. Rabiner, ``A tutorial on hidden markov models and selected applications
  in speech recognition,'' \emph{Proceedings of the IEEE}, vol.~77, no.~2, pp.
  257--286, 1989.

\bibitem{RaghunathanFDL16}
A.~Raghunathan, R.~Frostig, J.~Duchi, and P.~Liang, ``Estimation from indirect
  supervision with linear moments,'' in \emph{ICML}, 2016, pp. 2568--2577.

\bibitem{Rahman2011CRW}
A.~Rahman and V.~Ng, ``Coreference resolution with world knowledge,'' in
  \emph{ACL-HLT}, 2011, pp. 814--824.

\bibitem{RahmanN12}
------, ``Resolving complex cases of definite pronouns: The winograd schema
  challenge,'' in \emph{EMNLP-CoNLL}, 2012, pp. 777--789.

\bibitem{RainaBLPN07}
R.~Raina, A.~Battle, H.~Lee, B.~Packer, and A.~Y. Ng, ``Self-taught learning:
  transfer learning from unlabeled data,'' in \emph{ICML}, 2007, pp. 759--766.

\bibitem{RainaNK06}
R.~Raina, A.~Y. Ng, and D.~Koller, ``Constructing informative priors using
  transfer learning,'' in \emph{ICML}, 2006, pp. 713--720.

\bibitem{RanzatoPCL06}
M.~Ranzato, C.~S. Poultney, S.~Chopra, and Y.~LeCun, ``Efficient learning of
  sparse representations with an energy-based model,'' in \emph{NIPS}, 2006,
  pp. 1137--1144.

\bibitem{RatinovR09}
L.~Ratinov and D.~Roth, ``Design challenges and misconceptions in named entity
  recognition,'' in \emph{CoNLL}, 2009, pp. 147--155.

\bibitem{RatinovRDA11}
L.~Ratinov, D.~Roth, D.~Downey, and M.~Anderson, ``Local and global algorithms
  for disambiguation to wikipedia,'' in \emph{ACL}, 2011, pp. 1375--1384.

\bibitem{ReddyLS14}
S.~Reddy, M.~Lapata, and M.~Steedman, ``Large-scale semantic parsing without
  question-answer pairs,'' \emph{{TACL}}, vol.~2, pp. 377--392, 2014.

\bibitem{RiedelYM10}
S.~Riedel, L.~Yao, and A.~McCallum, ``Modeling relations and their mentions
  without labeled text,'' in \emph{{ECML} {PKDD}}, 2010, pp. 148--163.

\bibitem{RiedelYMM13}
S.~Riedel, L.~Yao, A.~McCallum, and B.~M. Marlin, ``Relation extraction with
  matrix factorization and universal schemas,'' in \emph{HLT-NAACL}, 2013, pp.
  74--84.

\bibitem{RocktaschelSR15}
T.~Rockt{\"{a}}schel, S.~Singh, and S.~Riedel, ``Injecting logical background
  knowledge into embeddings for relation extraction,'' in \emph{{NAACL}-{HLT}},
  2015, pp. 1119--1129.

\bibitem{RomeraParedesT15}
B.~Romera{-}Paredes and P.~H.~S. Torr, ``An embarrassingly simple approach to
  zero-shot learning,'' in \emph{ICML}, 2015, pp. 2152--2161.

\bibitem{RothJCC14}
D.~Roth, H.~Ji, M.~Chang, and T.~Cassidy, ``Wikification and beyond: The
  challenges of entity and concept grounding,'' in \emph{ACL, Tutorial
  Abstracts}, 2014, p.~7.

\bibitem{RussellAIM}
S.~Russell and P.~Norvig, \emph{Artificial Intelligence: A Modern Approach},
  3rd~ed.\hskip 1em plus 0.5em minus 0.4em\relax Prentice Hall Press, 2009.

\bibitem{SametFMM2005}
H.~Samet, \emph{Foundations of Multidimensional and Metric Data Structures (The
  Morgan Kaufmann Series in Computer Graphics and Geometric Modeling)}.\hskip
  1em plus 0.5em minus 0.4em\relax San Francisco, CA, USA: Morgan Kaufmann
  Publishers Inc., 2005.

\bibitem{saul06spectral}
L.~K. Saul, K.~Q. Weinberger, F.~Sha, J.~Ham, and D.~D. Lee, ``Spectral methods
  for dimensionality reduction,'' in \emph{Semi-Supervised Learning},
  O.~Chapelle, B.~Sch{\"o}lkopf, and A.~Zien, Eds.\hskip 1em plus 0.5em minus
  0.4em\relax {MIT} Press, 2006, ch.~16, pp. 293--30.

\bibitem{Schuhmacher2014}
M.~Schuhmacher and S.~P. Ponzetto, ``Knowledge-based graph document modeling,''
  in \emph{WSDM}, 2014, pp. 543--552.

\bibitem{ShenWH15}
W.~Shen, J.~Wang, and J.~Han, ``Entity linking with a knowledge base: Issues,
  techniques, and solutions,'' \emph{{IEEE} Trans. Knowl. Data Eng.}, vol.~27,
  no.~2, pp. 443--460, 2015.

\bibitem{ShiLZSY17}
C.~Shi, Y.~Li, J.~Zhang, Y.~Sun, and P.~S. Yu, ``A survey of heterogeneous
  information network analysis,'' \emph{{IEEE} Trans. Knowl. Data Eng.},
  vol.~29, no.~1, pp. 17--37, 2017.

\bibitem{SinghSPM11}
S.~Singh, A.~Subramanya, F.~C.~N. Pereira, and A.~McCallum, ``Large-scale
  cross-document coreference using distributed inference and hierarchical
  models,'' in \emph{ACL}, 2011, pp. 793--803.

\bibitem{SocherGMN13}
R.~Socher, M.~Ganjoo, C.~D. Manning, and A.~Y. Ng, ``Zero-shot learning through
  cross-modal transfer,'' in \emph{NIPS}, 2013, pp. 935--943.

\bibitem{SongDataless2017}
Y.~Song, S.~Mayhew, and D.~Roth, ``Cross-lingual dataless classification for
  languages with small wikipedia presence,'' \emph{arXiv preprint
  arXiv:1611.04122}, 2016.

\bibitem{song2013constrained}
Y.~Song, S.~Pan, S.~Liu, F.~Wei, M.~Zhou, and W.~Qian, ``Constrained text
  coclustering with supervised and unsupervised constraints,'' \emph{IEEE
  TKDE}, vol.~25, no.~6, pp. 1227--1239, 2013.

\bibitem{SongR14}
Y.~Song and D.~Roth, ``On dataless hierarchical text classification,'' in
  \emph{AAAI}, 2014, pp. 1579--1585.

\bibitem{SongR15}
------, ``Unsupervised sparse vector densification for short text similarity,''
  in \emph{{NAACL}-{HLT}}, 2015, pp. 1275--1280.

\bibitem{SUPR16}
Y.~Song, S.~Upadhyay, H.~Peng, and D.~Roth, ``Cross-lingual dataless
  classification for many languages,'' in \emph{IJCAI}, 2016, pp. 4223--4227.

\bibitem{Song14}
Y.~Song, H.~Wang, W.~Chen, and S.~Wang, ``Transfer understanding from head
  queries to tail queries,'' in \emph{CIKM}, 2014, pp. 1299--1308.

\bibitem{song2011short}
Y.~Song, H.~Wang, Z.~Wang, H.~Li, and W.~Chen, ``Short text conceptualization
  using a probabilistic knowledgebase,'' in \emph{IJCAI}, 2011, pp. 2330--2336.

\bibitem{song2015short}
Y.~Song, S.~Wang, and H.~Wang, ``Open domain short text conceptualization: A
  generative + descriptive modeling approach,'' in \emph{IJCAI}, 2015, pp.
  3820--3826.

\bibitem{Sorg2012}
P.~Sorg and P.~Cimiano, ``Exploiting wikipedia for cross-lingual and
  multilingual information retrieval,'' \emph{Data and Knowledge Engineering},
  vol.~74, pp. 26--45, 2012.

\bibitem{SrinivasanCS09}
H.~Srinivasan, J.~Chen, and R.~K. Srihari, ``Cross document person name
  disambiguation using entity profiles,'' in \emph{TAC}, 2009.

\bibitem{suchanek2007yago}
F.~M. Suchanek, G.~Kasneci, and G.~Weikum, ``Yago: a core of semantic
  knowledge,'' in \emph{WWW}, 2007, pp. 697--706.

\bibitem{SukhbaatarSWF15}
S.~Sukhbaatar, A.~Szlam, J.~Weston, and R.~Fergus, ``End-to-end memory
  networks,'' in \emph{NIPS}, 2015, pp. 2440--2448.

\bibitem{SunXW015}
X.~Sun, Y.~Xiao, H.~Wang, and W.~Wang, ``On conceptual labeling of a bag of
  words,'' in \emph{IJCAI}, 2015, pp. 1326--1332.

\bibitem{sun2012mining}
Y.~Sun and J.~Han, ``Mining heterogeneous information networks: principles and
  methodologies,'' \emph{Synthesis Lectures on Data Mining and Knowledge
  Discovery}, vol.~3, no.~2, pp. 1--159, 2012.

\bibitem{Yizhou11}
Y.~Sun, J.~Han, X.~Yan, P.~S. Yu, and T.~Wu, ``Pathsim: Meta path-based top-k
  similarity search in heterogeneous information networks.'' \emph{PVLDB}, pp.
  992--1003, 2011.

\bibitem{Sun2015hierarchical}
Y.~Sun, A.~Singla, D.~Fox, and A.~Krause, ``Building hierarchies of concepts
  via crowdsourcing,'' in \emph{IJCAI}, 2015, pp. 844--851.

\bibitem{SurdeanuTNM12}
M.~Surdeanu, J.~Tibshirani, R.~Nallapati, and C.~D. Manning, ``Multi-instance
  multi-label learning for relation extraction,'' in \emph{EMNLP-CoNLL}, 2012,
  pp. 455--465.

\bibitem{Surendran06}
A.~C. Surendran and S.~Sra, ``Incremental aspect models for mining document
  streams,'' in \emph{ECML/PKDD}, 2006, pp. 633--640.

\bibitem{SyedFJ08}
Z.~S. Syed, T.~Finin, and A.~Joshi, ``Wikipedia as an ontology for describing
  documents,'' in \emph{ICWSM}, 2008.

\bibitem{tenenbaum2011grow}
J.~B. Tenenbaum, C.~Kemp, T.~L. Griffiths, and N.~D. Goodman, ``How to grow a
  mind: Statistics, structure, and abstraction,'' \emph{Science}, vol. 331, no.
  6022, pp. 1279--1285, 2011.

\bibitem{TillmannN03}
C.~Tillmann and H.~Ney, ``Word reordering and a dynamic programming beam search
  algorithm for statistical machine translation,'' \emph{Computational
  Linguistics}, vol.~29, no.~1, pp. 97--133, 2003.

\bibitem{Torrey2009}
L.~Torrey and J.~Shavlik, ``Transfer learning,'' in \emph{Handbook of Research
  on Machine Learning Applications}, E.~Soria, J.~Martin, R.~Magdalena,
  M.~Martinez, and A.~Serrano, Eds.\hskip 1em plus 0.5em minus 0.4em\relax IGI
  Global, 2009.

\bibitem{ToutanovaCPPCG15}
K.~Toutanova, D.~Chen, P.~Pantel, H.~Poon, P.~Choudhury, and M.~Gamon,
  ``Representing text for joint embedding of text and knowledge bases,'' in
  \emph{EMNLP}, 2015, pp. 1499--1509.

\bibitem{TuMLCZ14}
K.~Tu, M.~Meng, M.~W. Lee, T.~E. Choe, and S.~C. Zhu, ``Joint video and text
  parsing for understanding events and answering queries,'' \emph{{IEEE}
  MultiMedia}, vol.~21, no.~2, pp. 42--70, 2014.

\bibitem{TurianRaBe10Clean}
J.~Turian, L.-A. Ratinov, and Y.~Bengio, ``Word representations: A simple and
  general method for semi-supervised learning,'' in \emph{ACL}, 2010, pp.
  384--394.

\bibitem{WainwrightJ08}
M.~J. Wainwright and M.~I. Jordan, ``Graphical models, exponential families,
  and variational inference,'' \emph{Foundations and Trends in Machine
  Learning}, vol.~1, no. 1-2, pp. 1--305, 2008.

\bibitem{Wang2015IWK}
C.~Wang, Y.~Song, A.~El-Kishky, D.~Roth, M.~Zhang, and J.~Han, ``Incorporating
  world knowledge to document clustering via heterogeneous information
  networks,'' in \emph{KDD}, 2015, pp. 1215--1224.

\bibitem{WangChenguangICDM2015}
C.~Wang, Y.~Song, H.~Li, M.~Zhang, and J.~Han, ``Knowsim: A document similarity
  measure on structured heterogeneous information networks,'' in \emph{ICDM},
  2015, pp. 1015--1020.

\bibitem{WangChenguangAAAI16}
------, ``Text classification with heterogeneous information network kernels,''
  in \emph{AAAI}, 2016, pp. 2130--2136.

\bibitem{WangChenguangSRZH16}
\BIBentryALTinterwordspacing
C.~Wang, Y.~Song, D.~Roth, M.~Zhang, and J.~Han, ``World knowledge as indirect
  supervision for document clustering,'' \emph{TKDD}, 2016. [Online].
  Available: \url{http://arxiv.org/abs/1608.00104}
\BIBentrySTDinterwordspacing

\bibitem{wang2014knowledge}
Z.~Wang, J.~Zhang, J.~Feng, and Z.~Chen, ``Knowledge graph and text jointly
  embedding,'' in \emph{EMNLP}, 2014, pp. 1591--1601.

\bibitem{Wang14}
Z.~Wang, F.~Wang, J.-R. Wen, and Z.~Li, ``Concept-based short text
  classification and ranking,'' in \emph{CIKM}, 2014, pp. 1069--1078.

\bibitem{WangZWMW15}
Z.~Wang, K.~Zhao, H.~Wang, X.~Meng, and J.~Wen, ``Query understanding through
  knowledge-based conceptualization,'' in \emph{IJCAI}, 2015, pp. 3264--3270.

\bibitem{WongM07}
Y.~W. Wong and R.~J. Mooney, ``Learning synchronous grammars for semantic
  parsing with lambda calculus,'' in \emph{ACL}, 2007.

\bibitem{wu2011taxonomy}
W.~Wu, H.~Li, H.~Wang, and K.~Q. Zhu, ``Probase: A probabilistic taxonomy for
  text understanding,'' in \emph{SIGMOD}, 2012, pp. 481--492.

\bibitem{XiangPPSY11}
E.~W. Xiang, S.~J. Pan, W.~Pan, J.~Su, and Q.~Yang, ``Source-selection-free
  transfer learning,'' in \emph{IJCAI}, 2011, pp. 2355--2360.

\bibitem{xu2014rc}
C.~Xu, Y.~Bai, J.~Bian, B.~Gao, G.~Wang, X.~Liu, and T.-Y. Liu, ``Rc-net: A
  general framework for incorporating knowledge into word representations,'' in
  \emph{CIKM}, 2014, pp. 1219--1228.

\bibitem{XuHZG13}
W.~Xu, R.~Hoffmann, L.~Zhao, and R.~Grishman, ``Filling knowledge base gaps for
  distant supervision of relation extraction,'' in \emph{ACL}, 2013, pp.
  665--670.

\bibitem{YangCXDY09}
Q.~Yang, Y.~Chen, G.~Xue, W.~Dai, and Y.~Yu, ``Heterogeneous transfer learning
  for image clustering via the socialweb,'' in \emph{ACL}, 2009, pp. 1--9.

\bibitem{YaoD14}
X.~Yao and B.~V. Durme, ``Information extraction over structured data: Question
  answering with freebase,'' in \emph{ACL}, 2014, pp. 956--966.

\bibitem{YazdaniH15}
M.~Yazdani and J.~Henderson, ``A model of zero-shot learning of spoken language
  understanding,'' in \emph{EMNLP}, 2015, pp. 244--249.

\bibitem{YihCHG15}
W.~Yih, M.~Chang, X.~He, and J.~Gao, ``Semantic parsing via staged query graph
  generation: Question answering with knowledge base,'' in \emph{ACL}, pp.
  1321--1331.

\bibitem{YihRMCS16}
W.~Yih, M.~Richardson, C.~Meek, M.~Chang, and J.~Suh, ``The value of semantic
  parse labeling for knowledge base question answering,'' in \emph{ACL}, 2016.

\bibitem{YosefHBSW11}
M.~A. Yosef, J.~Hoffart, I.~Bordino, M.~Spaniol, and G.~Weikum, ``{AIDA:} an
  online tool for accurate disambiguation of named entities in text and
  tables,'' \emph{{PVLDB}}, vol.~4, no.~12, pp. 1450--1453, 2011.

\bibitem{YuanGHDWZXLM15}
J.~Yuan, F.~Gao, Q.~Ho, W.~Dai, J.~Wei, X.~Zheng, E.~P. Xing, T.~Liu, and
  W.~Ma, ``Lightlda: Big topic models on modest computer clusters,'' in
  \emph{WWW}, 2015, pp. 1351--1361.

\bibitem{ZengLC015}
D.~Zeng, K.~Liu, Y.~Chen, and J.~Zhao, ``Distant supervision for relation
  extraction via piecewise convolutional neural networks,'' in \emph{EMNLP},
  2015, pp. 1753--1762.

\bibitem{ZhangYLXM16}
F.~Zhang, N.~J. Yuan, D.~Lian, X.~Xie, and W.~Ma, ``Collaborative knowledge
  base embedding for recommender systems,'' in \emph{SIGKDD}, 2016, pp.
  353--362.

\bibitem{ZhangSST11}
W.~Zhang, Y.~C. Sim, J.~Su, and C.~L. Tan, ``Entity linking with effective
  acronym expansion, instance selection, and topic modeling,'' in \emph{IJCAI},
  2011, pp. 1909--1914.

\bibitem{ZhengLHZ10}
Z.~Zheng, F.~Li, M.~Huang, and X.~Zhu, ``Learning to link entities with
  knowledge base,'' in \emph{HLT-NAACL}, 2010, pp. 483--491.

\bibitem{ZhongZWWC15}
H.~Zhong, J.~Zhang, Z.~Wang, H.~Wan, and Z.~Chen, ``Aligning knowledge and text
  embeddings by entity descriptions,'' in \emph{EMNLP}, 2015, pp. 267--272.

\bibitem{Zhu03}
S.~C. Zhu, ``Statistical modeling and conceptualization of visual patterns,''
  \emph{IEEE Transactions on Pattern Analysis and Machine Intelligence},
  vol.~25, no.~6, pp. 691--712, 2003.

\bibitem{ZhuRQK07}
X.~Zhu, T.~T. Rogers, R.~Qian, and C.~Kalish, ``Humans perform semi-supervised
  classification too,'' in \emph{AAAI}, 2007, pp. 864--870.

\end{thebibliography}
}

\end{document}